\pdfoutput=1

\documentclass[11pt]{article}

\usepackage{acl}

\usepackage{times}
\usepackage{latexsym}

\usepackage[T1]{fontenc}

\usepackage[utf8]{inputenc}

\usepackage{microtype}

\usepackage{inconsolata}

\usepackage{graphicx}

\title{Exploring Two-Phase Continual Instruction Fine-tuning for Multilingual Adaptation in Large Language Models}

\usepackage{centernot}
\usepackage{multicol}
\usepackage{multirow}
\usepackage{mathtools}
\usepackage{enumitem}
\usepackage{booktabs}
\usepackage{amsmath,amsfonts,amsthm}
\usepackage{adjustbox}
\usepackage{arydshln}
\usepackage{rotating}   
\usepackage{tcolorbox}

\newcommand{\malp}{\textsc{MultiAlpaca}}
\newcommand{\alp}{\textsc{Alpaca}}
\newcommand{\orca}{\textsc{OpenOrca}}
\newcommand{\morca}{\textsc{mOpenOrca}}
\newcommand{\llm}{\textsc{LLaMA-3-8B}}
\newcommand{\mis}{\textsc{Mistral-7B}}
\newcommand{\misi}{\textsc{Mistral-7B-Instruct}}
\newcommand{\llmi}{\textsc{LLaMA-3-8B-Instruct}}
\newtheorem{definition}{Definition}

\definecolor{pastelred}{rgb}{1.0, 0.41, 0.38}

\newcommand*\samethanks[1][\value{footnote}]{\footnotemark[#1]}
\author{
    Divyanshu Aggarwal\thanks{Joint First Author}\textsuperscript{1}, 
    Sankarshan Damle\samethanks\textsuperscript{1},
    Navin Goyal\textsuperscript{2}, 
    Satya Lokam\textsuperscript{1},
    Sunayana Sitaram\textsuperscript{2}
    \\
    \textsuperscript{1}Microsoft~
    \textsuperscript{2}Microsoft Research India \\
    \texttt{\{divyanshu.aggarwal,t-sandamle,navingo,satya.lokam,sunayana.sitaram\}@microsoft.com}
}

\begin{document}
\maketitle
\begin{abstract}
A key challenge for Large Language Models (LLMs) is improving their Multilingual instruction-following ability over time without deteriorating their ability in languages they already excel at, typically English. In this paper, we study a two-phase \emph{Continual Fine-tuning (CFT)} setup toward improving a model's Multilingual adaptability. Concretely, we consider a two-phase CFT process in which an English-only end-to-end instruction fine-tuned LLM (Phase 1) is sequentially fine-tuned on a multilingual instruction dataset (Phase 2). Across \mis\ and \llm\ and multiple dataset pairs, we show that instructional similarity between phases is critical: aligned datasets preserve or improve English while boosting multilingual ability, whereas misaligned datasets cause English degradation. We show that this degradation arises from representation shift during CFT, and that targeted mitigation strategies, including generative replay and heuristic-based layer freezing, reduce this shift and improve multilingual adaptation.
\end{abstract}

\section{Introduction}\label{sec::introduction}

%
\begin{figure}[t]
\centering
    \begin{minipage}{0.49\columnwidth}
    \centering
        \includegraphics[width=\columnwidth]{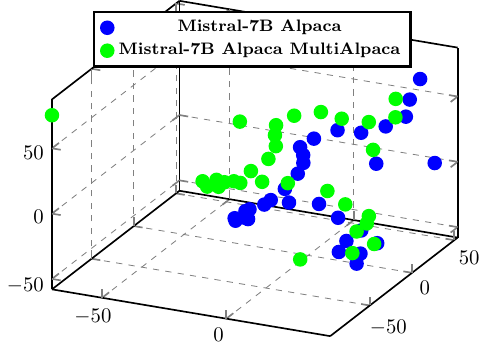}
    \end{minipage}
    \begin{minipage}{0.49\columnwidth}
    \centering
        \includegraphics[width=\columnwidth]{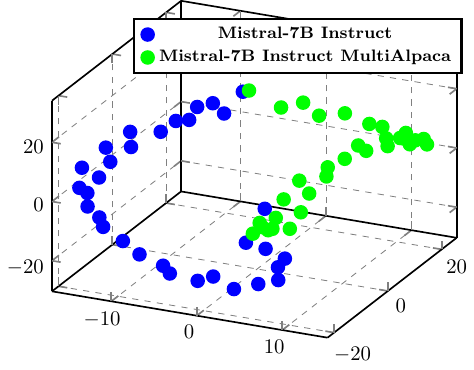}
    \end{minipage}
        \begin{minipage}{0.49\columnwidth}
    \centering
        \includegraphics[width=\columnwidth]{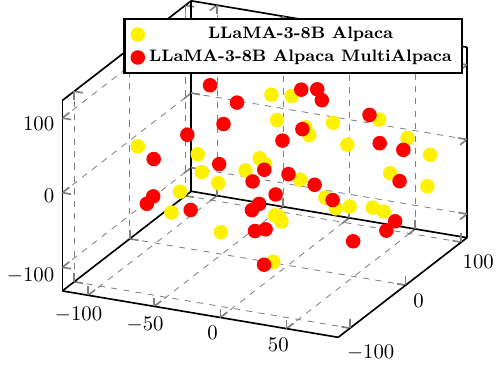}
    \end{minipage}
    \begin{minipage}{0.49\columnwidth}
    \centering
        \includegraphics[width=\columnwidth]{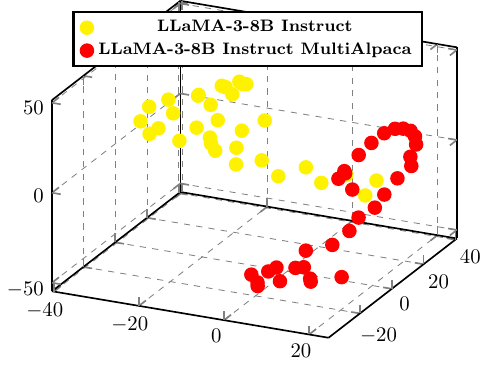}
    \end{minipage}
    \caption{Comparing \textit{t-SNEs}~\citep{vanDerMaaten2008} of the hidden activations for \mis\ and \llm\ during our two-phase Continual Fine-tuning (CFT) process. We prompt each model with examples from \textsc{MTBench}~\cite{zheng2024judging}, and visualize the similarity between the mean hidden activations, for each model layer. For datasets that encode "similar" instructions (\alp\ \& \malp), English ability does not decline (e.g., 3\% gain for \texttt{IFEval}). For non-similar datasets (Instruct \& \malp), English ability declines (e.g., 8\% decline for \texttt{IFEval}). Here, Phase 2 model representations do not align with Phase 1's; thus, suggesting greater model weight interference and a decline in English ability.   
    }\label{fig::tsne}
\end{figure}

The widespread adoption of Large Language Models (LLMs) has led to a growing multilingual user base~\cite{Shiyas2023}. However, ensuring strong performance across languages remains a fundamental challenge, with models consistently performing worse on low-resource languages spoken by millions of speakers worldwide~\citep{ahuja-etal-2023-mega,ahuja2024megaverse}. A key limitation is that both labeled and unlabeled training data are predominantly available in English and a few high-resource languages, while resources for other languages, especially low-resource ones, are scarce~\cite{shaham2024multilingual}. 

Training large models from scratch is computationally expensive, making \emph{fine-tuning} pre-trained LLMs the preferred approach for improving multilingual capabilities~\cite{lankford2023adaptmllm,nguyen2023seallms}. A common fine-tuning strategy is to train LLMs on an instruction-following dataset that contains a \textit{mixture} of languages. However, these datasets are often heavily skewed toward English and other high-resource languages, leading to a performance imbalance: models perform strongly in English but struggle with low-resource languages~\cite{dhamecha2021role,li2024improving,li2024languagerankermetricquantifying}. Further, prior works show that fine-tuning on a dataset that only contains non-English languages can hurt the model's performance on English due to \textit{catastrophic forgetting}~\cite{wu-etal-2025-continual,behrouz2025nested}, which is not desirable for most real-world scenarios due to the volume of English queries~\cite{ta2023languagegaps}. Ideally, we want the same model to be proficient in both English and other languages to avoid the costs of maintaining multiple models. We refer to an LLM’s proficiency in English as its \emph{English Ability} (EA), and its effectiveness across other languages its \emph{Multilingual Ability} (MA). In this work, we aim to improve an LLM's MA while maintaining or improving its EA.

\subsection{Our Approach}
To bridge the gap between EA and MA, we introduce a \textit{two-phase Continual Fine-tuning} (CFT) setup. We fine-tune a pre-trained LLM on an English instruction dataset in Phase 1 and then fine-tune it on a similarly-sized Multilingual dataset in Phase 2. In Phase 1, we use \alp~\cite{alpaca} and \orca~\cite{OpenOrca}, and in Phase 2 we use \malp~\cite{multialpaca} and \morca~(\S\ref{subsec::exp_setup}).
\alp\ and \orca\ provide high-quality English instruction data, while \malp\ and \morca\ are their multilingual counterparts, ensuring consistency in instruction style across phases.
To compare the efficacy of our two-phase CFT setup, we compare it with a straightforward single-phase setup where the LLM is fine-tuned on the \textit{mixture} of both the instruction tuning datasets.

We focus on two open-source models, \llm\ and \mis\ as base models for our experiments. We also use fine-tuned versions of them, \llmi\ and \misi, as off-the-shelf Phase 1 English fine-tuned models\footnotemark. We quantify a model's English Ability (EA) based on its performance on four English datasets: (i) Two datasets that measure instruction following capabilities (i.e., \texttt{IFEval}~\cite{zhou2023instruction} and \texttt{Alpaca Eval}~\cite{li2023alpacaeval}) and (ii) two that measure reasoning abilities (i.e., MMLU~\cite{hendryckstest2021} and \texttt{HellaSwag}~\cite{zellers2019hellaswag}). Likewise, we quantify a model's Multilingual Ability (MA) based on its performance on (i) two question-answering tasks (i.e., \texttt{MLQA}~\cite{lewis2019mlqa} and \texttt{XQuAD}~\cite{Artetxe:etal:2019}) and (ii) \texttt{XLSUM}~\cite{hasan-etal-2021-xl}, a summarization task.  

\footnotetext{\llm's pre-training data was 5\% multilingual, but \llmi\ is primarily non-multilingual~\citep{llama}.}

\subsection{Our Contributions}

\paragraph{CFT Outperforms Mixture.} We first observe that models trained using our two-phase CFT setup perform better than the single-phase "dataset mixture" setup (Tables~\ref{tab:results_main},~\ref{tab:results_main_MA};~\S\ref{subsec::ta_la_results}). Moreover, our two-phase CFT setup overall results in a better model for all languages, including English, for the same number of training steps. The two-phase CFT pipeline also provides more flexibility than training on a mixture of datasets, with the possibility of extending our approach to multi-phase fine-tuning, especially when data from earlier phases might not be available. 

\paragraph{Forgetting vs. Dataset Similarity.} As mentioned earlier, fine-tuning with multilingual datasets to enhance a model's multilingual ability can lead to a decline in its English ability due to catastrophic forgetting~\citep{mukhoti2023fine,winata2023overcoming}. We investigate the factors that may lead to such forgetting by computing the similarity of English and Multilingual Instruction Fine-tuning (IFT) datasets. We observe that when English and multilingual datasets have instructions that are not similar, there is a decline in the Phase 2 model's performance in English. On the other hand, when Phase 1 and Phase 2 datasets encode similar instructions, the Phase 2 model's performance in English improves (refer to Figure~\ref{fig::tsne}). To quantify the similarity of these phase-wise datasets, we introduce two metrics based on language-agnostic embeddings and model representations. We show that our quantification correlates with the decline in English ability (Tables~\ref{tab:des},~\ref{tab:mps};~\S\ref{sec:dataset_diff}). 

\paragraph{Mitigating Forgetting.}  We study the efficacy of two tailored variants of existing CFT strategies to mitigate the decline in EA after Phase 2 fine-tuning, while boosting MA. The first strategy is distribution replay. Here, we look at \emph{generative replay}, i.e., using instructions from a similar English counterpart of the Phase 2 dataset to generate replay data using the Phase 1 model. We also try \emph{english replay} which acts as language replay by utilizing existing English parallel data from the Phase 2 distribution.
The second strategy employs \emph{layer freezing}. Our heuristic selects specific layers for freezing during Phase 2 fine-tuning based on the weight differences between the Base and Phase 1 models. We also explore Spectrum~\cite{hartford2024spectrum} as an alternative heuristic.
We study the gains in EA and MA of these strategies compared to specific baselines (Table~\ref{tab:results_interventions};~\S \ref{sec:mitigation}). To the best of our knowledge, we are the first to explore the effectiveness of CFT on LLMs with multilingual instruction datasets.


\section{Related Work}\label{sec::rw} 

\paragraph{Continual Learning in LLMs~\cite{chen2026continual}.}   In general, continual learning in LLMs can be broadly categorized into (i) continual pre-training (CPT) and (ii) continual fine-tuning (CFT). In CPT, the LLMs are continuously pre-trained to adapt to new domains or tasks by continuously updating them with new data alongside the existing data~\cite{shi2024continual}. CPT builds on the existing LLM's knowledge and is more computationally efficient than retraining an LLM using the current and old pre-training data~\cite{gupta2023continual}. CPT is employed when distributional shifts occur (i) over time~\cite{amba2021dynamic,jang2022temporalwiki,jangtowards}, (ii) across languages~\cite{jin2022lifelong,fujii2024continual,blevins2024breaking} or (iii) across domains~\cite{KeSLKK023,gong2022continual,xie2023efficient}. 

 On the other hand, CFT involves training the LLM on successive downstream tasks with varying data distribution or time shifts~\cite{shi2024continual}. CFT comprises fine-tuning for different tasks~\cite{carrion2022few,guan-etal-2025-multi}, instruction-tuning~\cite{cahyawijaya2023instructalign,kang2025self},  model refinement/editing~\cite{zhang2023copf} and alignment~\cite{suhr2023}. Recent literature also focuses on using CFT to assist the LLM to learn new languages~\cite{praharaj2023multilingual,pfeiffer-etal-2022-lifting,badola-etal-2023-parameter,singh2024three}.

\paragraph{CFT: Enhancing LLMs Multilingual Abilities.} \citeauthor{cahyawijaya2023instructalign}~\shortcite{cahyawijaya2023instructalign} propose InstructAlign which uses cross-lingual alignment and episodic replay to align an LLM's pre-trained languages to unseen languages but requires parallel data and previous task data. \citeauthor{shaham2024multilingual}~\shortcite{shaham2024multilingual} introduces multilinguality during the first instruction fine-tuning phase which improves an LLM's instruction following capability across languages. \citeauthor{he2023continual}~\shortcite{he2023continual} show catastrophic forgetting during CFT and use techniques such as joint fine-tuning and model regularization to mitigate it. However, these techniques are computationally expensive or require access to previous task data.

\paragraph{Multilingual Adaptation.}
This set of works looks at language and task adaption by adjusting the model to understand new languages and enhancing its performance on specific tasks through fine-tuning, respectively~\cite{chen2024improvinglanguageplasticitypretraining, zhao2024adamergexcrosslingualtransferlarge, pfeiffer2020madxadapterbasedframeworkmultitask}. For instance, \citet{chen2024improvinglanguageplasticitypretraining} perform task adaption by fine-tuning the model on downstream task data. For language adaption, they fine-tune only the token embedding layer, helping the model learn specific lexical meanings of new languages. Language and english ability are either trained in parallel or sequentially. 
However, in this paper, we try to incorporate multilingual ability in models with the constraint that they may have already learned english ability (e.g., \misi). To the best of our knowledge, this is a first attempt at studying the effect of task and language self-instruct datasets on an LLM's multilingual ability through CFT.

\section{Two-phase Continual Fine-tuning Setup}

\label{sec::exp_setup}

When instruction fine-tuning LLMs, the most natural method is to fine-tune on a "dataset mixture" containing English and Multilingual data~\cite{workshop2023bloom176bparameteropenaccessmultilingual}.  
However, fine-tuning on all languages simultaneously may introduce performance bias where the model performs better in English (and other high resource languages)~\cite{dhamecha2021role,li2024improving,li2024languagerankermetricquantifying}\footnotemark.

\footnotetext{In \S\ref{subsec::ta_la_results}, we compare dataset mixture to CFT.}

\paragraph{Continual Fine-tuning (CFT).} To improve the multilingual performance of pre-trained LLMs, we introduce the following two-phase CFT process. 
\begin{tcolorbox}[colframe=blue!75!white, colback=blue!5!white, coltitle=white, sharp corners=all, boxrule=0.8mm, title=Two-Phase CFT Process]
    \begin{itemize}[leftmargin=*]
        \item \textbf{Phase 1:} Fine-tune a base LLM end-to-end on an English instruction dataset. Phase 1 aims to teach the LLM \textit{English Instruction Following Ability}, which we refer to as \emph{English Ability} (EA).
        \item \textbf{Phase 2:} Take the fine-tuned LLM from Phase 1 and further fine-tune it end-to-end on a Multilingual instruction dataset. Phase 2 focuses on enhancing the LLM's \emph{Multilingual Ability} (MA), using a dataset with multiple languages and fewer data points per language.
    \end{itemize}
\end{tcolorbox}

\paragraph{Challenges.} The primary challenge in our two-phase CFT process is that the LLM's Multilingual Ability must not come at the cost of its English Ability. We impose \textit{two additional constraints} based on real-world scenarios. First, in Phase 2, we cannot re-use Phase 1's dataset. Often instruction fine-tuned LLMs are available without their corresponding datasets (e.g., \misi~\cite{mistral}). Second, in Phase 2, we cannot use the weights of the Phase 1 model during training, as saving both old and new set of parameters on the GPU for training would be computationally expensive.


%
\begin{table*}[t!]
    \centering
    \begin{adjustbox}{max width=\textwidth}
    \begin{tabular}{ccccccccccccccc}
    \toprule
    \multicolumn{15}{c}{\textbf{Two-phase Continual Fine-tuning}} \\
    \midrule
    \multirow{2}{*}{\textbf{Model}}     & \textbf{Phase 1 (P1)} & \textbf{Phase 2 (P2)} & \multicolumn{2}{c}{\texttt{IFEval} $(\uparrow)$}  & \multicolumn{2}{c}{\texttt{Alpaca Eval} $(\uparrow)$} &  \multicolumn{2}{c}{\texttt{MMLU} $(\uparrow)$} & \multicolumn{2}{c}{\texttt{HellaSwag} $(\uparrow)$} & \multicolumn{2}{c}{\texttt{XLSUM\_en} $(\uparrow)$} & \multicolumn{2}{c}{\textbf{Average}} \\
    & \textbf{Dataset} & \textbf{Dataset} & \textbf{P1} & \textbf{P2}  & \textbf{P1} &  \textbf{P2} & \textbf{P1} & \textbf{P2}  & \textbf{P1} & \textbf{P2} & \textbf{P1} & \textbf{P2} & \textbf{P1} & \textbf{P2} \\ 
    \midrule
    \multirow{2}{*}{\mis} & \alp & & 0.364  & \textcolor{green!80!black}{0.395}   & 0.12   & \textcolor{green!80!black}{0.16} &  0.552  &  \textcolor{green!80!black}{0.573}  &      {0.581} & \textcolor{green!80!black}{0.616} & 0.10 & \textcolor{green!80!black}{0.11} &  0.343 & \textcolor{green!80!black}{0.371}  \\  
    & Instruct & \textsc{Multi} & 0.550  & \textcolor{pastelred}{0.462}  &  {0.35}  & \textcolor{pastelred}{0.15} & {0.575}   & \textcolor{pastelred}{0.533}  &  0.641  & \textcolor{pastelred}{0.416} &  0.13 & \textcolor{pastelred}{0.10} & 0.449 & \textcolor{pastelred}{0.332}    \\ 
    \multirow{2}{*}{\llm} & \alp & \textsc{Alpaca}& 0.277  &  \textcolor{green!80!black}{0.326}  &  0.10  &  \textcolor{green!80!black}{0.11} & 0.231   &  \textcolor{green!80!black}{0.242}  &  0.556 &  \textcolor{green!80!black}{0.567} &   0.07  &  \textcolor{green!80!black}{0.08} &  0.247  &  \textcolor{green!80!black}{0.265}  \\
    & Instruct &  &  {0.735} &  \textcolor{pastelred}{0.182} &  0.14  & \textcolor{pastelred}{0.10} &  0.340  & \textcolor{pastelred}{0.239}  & 0.533  &  \textcolor{pastelred}{0.278} &  0.11 & \textcolor{pastelred}{0.09} & 0.372  & \textcolor{pastelred}{0.178}  \\ 
    \bottomrule
    \multicolumn{15}{c}{\textbf{Dataset Mixture}} \\
    \midrule
    \textbf{Model} & \multicolumn{2}{c}{\textbf{Dataset Mixture}} & \multicolumn{2}{c}{\texttt{IFEval} $(\uparrow)$} & \multicolumn{2}{c}{\texttt{Alpaca Eval} $(\uparrow)$} & \multicolumn{2}{c}{\texttt{MMLU} $(\uparrow)$} & \multicolumn{2}{c}{\texttt{HellaSwag} $(\uparrow)$} & \multicolumn{2}{c}{\texttt{XLSUM\_en} $(\uparrow)$} & \multicolumn{2}{c}{\textbf{Average}} \\
    \midrule
    \mis & \multicolumn{2}{c}{\alp\ \malp} & \multicolumn{2}{c}{0.394} & \multicolumn{2}{c}{0.23} & \multicolumn{2}{c}{0.538}  & \multicolumn{2}{c}{0.602}  & \multicolumn{2}{c}{0.09} &  \multicolumn{2}{c}{0.371} \\ 
    \llm & \multicolumn{2}{c}{\alp\ \malp} & \multicolumn{2}{c}{0.363} & \multicolumn{2}{c}{0.07} & \multicolumn{2}{c}{0.598}  & \multicolumn{2}{c}{0.602}  & \multicolumn{2}{c}{0.04}  & \multicolumn{2}{c}{0.335} \\
    \bottomrule
    \end{tabular}
    \end{adjustbox}
    \caption{\textbf{English Ability results for two-phase Continual Fine-tuning (CFT).} When the phase-wise datasets are similar (Definition~\ref{def::des} and Definition~\ref{def::mps}), English Ability post Phase 2 (P2) fine-tuning \textit{consistently} improves (denoted with \textcolor{green!80!black}{green}). When the phase-wise datasets are not similar, we see a \textit{significant} decline in English Ability post Phase 2 (P2) fine-tuning (denote with \textcolor{pastelred}{red}). We also provide numbers for dataset mixture -- when the models are fine-tuned simultaneously on the Phase 1 and Phase 2 datasets. 
    }
    \label{tab:results_main}
\end{table*}



\section{Evaluating English \& Multilingual Ability for Multilingual CFT\label{sec::langadapt_cft}}

\subsection{Experiment Setup \& Evaluation Tasks\label{subsec::exp_setup}}
\paragraph{Fine-tuning Models.} We continually fine-tune open-source \mis~\cite{mistral} and \llm~\cite{llama} LLMs for multilingual adaptation.

\paragraph{Fine-tuning Datasets.} For our phase-wise datasets, we use the open-source \alp~\cite{alpaca}, \malp~\cite{multialpaca}, and \orca~\cite{OpenOrca} datasets. \alp\ is a self-instruct English-only dataset. \malp\ is a multilingual dataset created by translating \alp's seed tasks to 11 languages and using GPT-3.5-Turbo for response collection. The languages are in equal proportions and are ``French'', ``Arabic'', ``German'', ``Spanish'', ``Indonesian'', ``Japanese'', ``Korean'', ``Portuguese'', ``Russian'', ``Thai'', and ``Vietnamese''. The appendix (\S\ref{app::fine-tuning-datasets}) describes \orca\ and \morca.

\paragraph{Fine-tuning Technique.}
We perform full fine-tuning in \texttt{bf16} precision to study the impact of multilingual post-training in Phase~2 on English ability. This setting allows us to fully leverage model capacity, which may not be attainable under parameter-efficient fine-tuning methods~\cite{aggarwal2024maple,panda2024lotteryticketadaptationmitigating}. To mitigate the resulting degradation in English performance, \S\ref{sec:mitigation} introduces a heuristic-based layer freezing strategy that selectively freezes a subset of layers while fine-tuning the remainder. All experiments are conducted using \emph{Axolotl}\footnote{\url{github.com/axolotl-ai-cloud/axolotl/}}, an open-source framework for large-scale LLM fine-tuning.

\paragraph{Evaluation Tasks.} To quantify an LLM's english ability, we evaluate Phase 1 and Phase 2 models on two instruction-following tasks (i) \texttt{IFEval}~\cite{zhou2023instruction} and (ii) \texttt{Alpaca Eval}~\cite{li2023alpacaeval}, (iii) \texttt{\texttt{MMLU}}~\cite{hendryckstest2021} for problem-solving, (iv) \texttt{\texttt{HellaSwag}}~\cite{zellers2019hellaswag} for commonsense reasoning ability, and (v) \texttt{XLSUM\_en}~\cite{hasan-etal-2021-xl} for summarization. To quantify an LLM's multilingual ability, we evaluate our fine-tuned models on three benchmark datasets comprising two multilingual generative tasks: question answering (\texttt{\texttt{MLQA}} \cite{lewis2019mlqa} \& \texttt{\texttt{XQuAD}} \cite{Artetxe:etal:2019}), instruction-following (\texttt{GMMLU}~\cite{singh2024global}), and summarization (\texttt{XLSUM}  \cite{hasan-etal-2021-xl}). 

Our evaluation suite spans diverse linguistic phenomena and reasoning skills to comprehensively assess language understanding, generation, and problem-solving. It also incorporates parallel benchmarks for English and multilingual ability, enabling direct cross-lingual comparison. Further details are provided in \S \ref{app: eval details}.

 To evaluate our models on EA and MA, we use \emph{LM-Evaluation-Harness}\footnote{\url{github.com/EleutherAI/lm-evaluation-harness}}, which is a unified framework for zero/few-shot evaluations of LLMs. For both English and multilingual ability, we use \textbf{zero-shot} evaluation. For additional details on the training setup, code, and evaluation tasks, refer to \S \ref{app::training}. 

\subsection{Results\label{subsec::ta_la_results}}

%
\begin{table*}[t!]
    \centering
    \begin{adjustbox}{max width=0.95\textwidth}
    \begin{tabular}{ccccccccccccc}
    \toprule
    \multicolumn{13}{c}{\textbf{Two-phase Continual Fine-tuning}} \\
    \toprule
    \multirow{2}{*}{\textbf{Model}}     & \textbf{Phase 1 (P1)} & \textbf{Phase 2 (P2)} & \multicolumn{2}{c}{\texttt{MLQA} $(\uparrow)$}  & \multicolumn{2}{c}{\texttt{XLSUM} $(\uparrow)$} &  \multicolumn{2}{c}{\texttt{XQuAD} $(\uparrow)$} & \multicolumn{2}{c}{\texttt{GMMLU} $(\uparrow)$} & \multicolumn{2}{c}{\textbf{Average}} \\
    & \textbf{Dataset} & \textbf{Dataset} & \textbf{P1} & \textbf{P2}  & \textbf{P1} &  \textbf{P2} & \textbf{P1} & \textbf{P2} & \textbf{P1} & \textbf{P2}  & \textbf{P1} & \textbf{P2}   \\
    \midrule
    \multirow{2}{*}{\mis} & \alp &  & 0.229  & \textcolor{green!80!black}{0.288}  &  0.012  & \textcolor{green!80!black}{0.060} & 0.290 & \textcolor{green!80!black}{0.602}  & 0.42 &  \textcolor{pastelred}{0.40} & 0.238 & \textcolor{green!80!black}{0.338}  \\  
    & Instruct & \textsc{Multi} & 0.246 & \textcolor{green!80!black}{0.307} & 0.012   & \textcolor{green!80!black}{0.033}  &  0.351 &   \textcolor{green!80!black}{0.436} & 0.44 & \textcolor{pastelred}{0.43} & 0.262 &   \textcolor{green!80!black}{0.302}\\  
    \multirow{2}{*}{\llm} & \alp & \textsc{Alpaca} & 0.438  &  \textcolor{green!80!black}{0.597}  &   0.033 &  \textcolor{green!80!black}{0.034} &  0.586 & \textcolor{green!80!black}{0.737}  & 0.53 & \textcolor{pastelred}{0.34} & 0.397 & \textcolor{green!80!black}{0.427} \\  
    & Instruct &   &  {0.609} &  \textcolor{pastelred}{0.321} & {0.048}  & \textcolor{pastelred}{0.027} & {0.712}   &  \textcolor{pastelred}{0.417}  & 0.25 & \textcolor{green!80!black}{0.44}  & 0.405 & \textcolor{pastelred}{0.301} \\  
    \bottomrule
    \multicolumn{13}{c}{\textbf{Dataset Mixture}} \\
    \midrule
    \textbf{Model} & \multicolumn{2}{c}{\textbf{Dataset Mixture}} & \multicolumn{2}{c}{\texttt{MLQA} $(\uparrow)$} & \multicolumn{2}{c}{\texttt{XLSUM} $(\uparrow)$} & \multicolumn{2}{c}{\texttt{XQuAD} $(\uparrow)$} & \multicolumn{2}{c}{\texttt{GMMLU} $(\uparrow)$}  & \multicolumn{2}{c}{\textbf{Average}} \\
    \midrule
    \mis & \multicolumn{2}{c}{\alp\ \malp} & \multicolumn{2}{c}{0.406} & \multicolumn{2}{c}{0.079} & \multicolumn{2}{c}{0.217}  &  \multicolumn{2}{c}{0.41} & \multicolumn{2}{c}{0.278} \\ 
    \llm & \multicolumn{2}{c}{\alp\ \malp} & \multicolumn{2}{c}{0.480} & \multicolumn{2}{c}{0.040} & \multicolumn{2}{c}{0.139}  & \multicolumn{2}{c}{0.50} &  \multicolumn{2}{c}{0.289} \\
    \bottomrule
    \end{tabular}
    \end{adjustbox}
    \caption{\textbf{Multilingual Ability results for two-phase Continual Fine-tuning (CFT).} With \textcolor{green!80!black}{green}, we denote an improvement in Multilingual Ability post Phase 2 fine-tuning. Likewise, we denote a decline in Multilingual Ability with \textcolor{pastelred}{red}. For \texttt{MLQA} and \texttt{XQUAD} we use F1 abstractive score, while for \texttt{XLSUM} we use {ROUGE}~\cite{lin-2004-rouge} score.  We also provide numbers for dataset mixture -- when the models are fine-tuned simultaneously on the Phase 1 and Phase 2 datasets.
    }
    \label{tab:results_main_MA}
\end{table*}

We compare the English and Multilingual ability of \mis\ and \llm\ continually fine-tuned models on different phase-wise datasets\footnote{When it is clear from the context, we use ``Instruct'' to denote the dataset used in Phase 1 to instruction fine-tune \misi\ or \llmi.}. Table~\ref{tab:results_main} presents the results for English Ability (EA), while Table~\ref{tab:results_main_MA} presents the results for Multilingual Ability (MA). Table~\ref{tab:results_main_MA} reports the average score across languages. We provide language-specific scores and results when the phases are reversed (e.g., \malp-\alp) in \S\ref{app::more_Evals} and \S\ref{app::reverse_order}.

\paragraph{Comparison with Mixture.} 
From Tables~\ref{tab:results_main}~\&~\ref{tab:results_main_MA}: for Mixture, the mean of EA and MA scores for \mis\ fine-tuned on \alp-\malp\ is 0.325, and 0.312 for \llm. The corresponding two-phase mean score is 0.355 for \mis\ and 0.346 for \llm. Our two-phase CFT setup is more effective than Mixture, for approximately the same number of training steps.

\paragraph{Discussion.} From Table~\ref{tab:results_main}, for phase-wise datasets like Instruct and \malp, the performance of the Phase 2 models trained on them declines for English. This decline occurs when they are {continually} fine-tuned on multilingual data in Phase 2. However, we see a jump in \mis's multilingual ability for the multilingual generative tasks (Table~\ref{tab:results_main_MA}). That is, Phase 2 models fine-tuned on multilingual datasets show forgetting in English. However, for phase-wise datasets like \alp\ followed by \malp, we see that Phase 2 models do not show a decline in English ability (Table~\ref{tab:results_main}). We also see a gain in these models' multilingual ability (Table~\ref{tab:results_main_MA}).

\paragraph{Ablations.} In Tables~\ref{tab:results_main_ablation}~\&~\ref{tab:results_main_LA_supp}~(\S \ref{app::more_Evals}), we present results for \orca-\morca\ phase-wise datasets. First, the "dataset mixture" again performs worse on average than CFT: 0.186 vs. 0.372 for \mis\ and 0.217 vs. 0.366 for \llm. Second, for \mis, the average English ability of the Phase 2 model (over Phase 1's \mis-\orca) marginally improves: 0.413 from 0.407. Whereas, for \misi, the average decline in English ability is significant: 0.30 from 0.450. Likewise, for \llm, the average English ability for \llm\ \orca\ \morca\ sees an increase to 0.356 from 0.337. In contrast, for Instruct-\morca, the English ability significantly drops, from 0.372 to 0.138.

\paragraph{Observation.} With Table~\ref{tab:results_main}, we see that our two-phase CFT setup for multilingual adaptation shows an interesting trend: for certain pairs of phase-wise datasets (e.g., \alp\ \& \malp), the LLM after Phase 2 sees an improvement in the English ability (computed on English evaluation tasks). We notice that phase-wise datasets like \alp\ and \malp\ have the same seed prompts. Alternately, the two datasets \textit{encode the same instructions in different languages}. We hypothesize an LLM fine-tuned on either of these datasets learns the same instructions, and therefore, the second phase of CFT leads to lesser interference in the representation space. That is, an LLM continually fine-tuned on \alp\ \& \malp\ preserves its English ability across phases. We next define two metrics that aim to quantify the instruction-specific similarity of two datasets.

\subsection{Similarity of Phase-wise Datasets}
\label{sec:dataset_diff}

%
\begin{table}[t]
    \centering
    \begin{adjustbox}{max width=\columnwidth}
    \begin{tabular}{ccc}
    \toprule
    \textbf{Phase 1 Dataset} & \textbf{Phase 2 Dataset} & \textbf{DES ($\uparrow$)}  \\
    \midrule
    \multirow{2}*{\alp} & \malp &  0.924   \\  
     & \morca & 0.792 \\ 
     \hdashline
    \multirow{2}*{\orca} & \morca  &   0.953  \\  
     & \malp  & 0.774 \\
     \hdashline
    \mis\ Instruct$^\ddagger$ & \malp & 0.746 \\
    \bottomrule
    \multicolumn{3}{l}{$^\ddagger$: Prepared using model responses on \textsc{MTBench}~\cite{zheng2024judging}}
    \end{tabular}
    \end{adjustbox}
    \caption{Quantifying Phase-wise Dataset Similarity using DES: higher the score, greater the dataset similarity. }
    \label{tab:des}
\end{table}

\paragraph{Dataset Embedding Similarity (DES).} To quantify whether two datasets are similar\footnotemark, we define DES that computes a similarity score using the dot product of the average representations (embeddings) generated by a language-agnostic model. 
\begin{definition}[Dataset Embedding Similarity (DES)]\label{def::des}
    Given a language-agnostic text embedding model $\Theta$, and any pair of datasets $D_1$ and $D_2$, let DES be the function $f_\textsf{DES}:D \times D \rightarrow [0,1]$
    \begin{equation*}
        f_\textsf{DES}(D_1,D_2;\Theta) = \langle \mathbf{E}_\Theta(D_1), \mathbf{E}_\Theta(D_2) \rangle
    \end{equation*}
    Here, $\mathbf{E}_\Theta(D_i)\in\mathbb{R}^d,~\forall i\in\{1,2\}$ is the normalized mean embedding across samples in $D_i$.
\end{definition}
\footnotetext{The CL-ML literature often defines task similarity via permutation tasks, emphasizing input-output transformations~\cite{goldfarbjoint}. Whereas, we consider semantic and structural similarity in natural language instructions.}

Higher the DES score, more similar the embedding, i.e., greater similarity between $D_1$ and $D_2$. For $\Theta$, we use the language-agnostic sentence-tokenizer LaBSE~\cite{feng2020languageagnostic}. We compute DES by encoding 500 random samples from \alp, \malp, \orca, and \morca, and measure $f_\textsf{DES}$ for each pair. Table \ref{tab:des} presents the numbers. For dataset pairs with similar datasets, we see a high DES score and relatively low scores for dissimilar datasets. DES captures the (pair-wise) variation in instruction similarity of these datasets.

%
\begin{table}[t]
    \centering
    \begin{adjustbox}{max width=0.75\columnwidth}
    \begin{tabular}{cc}
    \toprule
      \textbf{Dataset} $D_2$  & \textbf{Model Parameter Difference ($\downarrow$)}  \\
    \midrule
     \alp  & 0.29 \\
      Instruct &  1.00 \\
     \orca &   0.55 \\
     \bottomrule
    \end{tabular}
    \end{adjustbox}
    \caption{Quantifying Phase-wise Similarity using MPD: lower the score, greater the dataset similarity. Here, we fix \malp\ as $D_1$ and $\theta_B$ as \mis. 
    }
    \label{tab:mps}
\end{table}

\paragraph{Model Parameter Difference (MPD).} Another method of quantifying the similarity of instructions for two datasets $D_1$ and $D_2$ is to compute the difference between the parameters of models $\Theta_1$ (fine-tuned on $D_1$) and $\Theta_2$ (fine-tuned on $D_2$). Geometrically, the difference of the parameters captures the representation shift of $\Theta_2$ in the space defined by $\Theta_1$. If $D_1$ \& $D_2$ encode the same datasets, the combined shift by $\Theta_2$ should be relatively lower, compared to the shift if $D_1$ \& $D_2$ encode different intstructions. Formally,

\begin{definition}[Model Parameter Difference (MPD)]\label{def::mps}
    Given any two models $\Theta_1$ and $\Theta_2$ fine-tuned on self-instruct datasets $D_1$ and $D_2$ respectively, from the same base model $\Theta_B$, let MPD be the function $f_\textsf{MPD}:\Theta \times \Theta \rightarrow \mathbb{R}_{\geq 0}$ s.t.
        \begin{equation*}
            f_\textsf{MPD}(\Theta_1,\Theta_2;\Theta_B) = \frac{1}{n} \sum_{i=1}^{n} \| \mathbf{w}(\Theta_{1,i}) - \mathbf{w}(\Theta_{2,i}) \|_2
        \end{equation*}
    Here, $\mathbf{w}(\Theta_{j,i}),~\forall j \in \{1,2\}$ is $\Theta_j$'s $i\textsuperscript{th}$ parameter.
\end{definition}

The smaller the MPD score, the closer the fine-tuned models are in the parameter space. Fixing \mis\ as the base model $\Theta_B$, and $D_1$ as \malp, we vary $D_2$ as one of \alp,~\orca, and \morca, and observe the corresponding MPD scores. We normalize the MPD scores with the maximum observed score across all three models for a fair comparison (see Table \ref{tab:mps}). MPD shows a similar trend to DES: for \alp\ and \malp, the scores are lower, highlighting the similarities in the datasets in the parameter space. We see relatively higher scores for the other pair of models, implying a difference in the dataset pairs.

\subsection{Visualizing Decline in English Ability\label{subsec::vs_decline}}
\paragraph{Setup.} To explain the effect of similar phase-wise data sets on an LLM's EA, we look at model representations when parsing English. We feed \textsc{MTBench}~\cite{zheng2024judging} to the models, a widely-used English benchmark for generalized instruction-following evaluation, and visualize the similarity between the mean hidden activations for each model layer. For the analysis, given an LLM $\Theta$ with $l$ layers, let $X_\Theta \in \mathbb{R}^{l\times d}$ be the mean hidden activations, across $n$ samples from \textsc{MTBench}.

\paragraph{t-SNE Visualization.} Figure~\ref{fig::tsne} depicts t-SNEs~\cite{vanDerMaaten2008} for $X_{\mis}$ and $X_{\llm}$ when these are continually fine-tuned on (i) \alp\ \& \malp\ and (ii) Instruct \& \malp. We observe that for similar phase-wise datasets, the model before and after Phase 2 produces similar hidden activations. Contrarily, for non-similar phase-wise datasets, the hidden activations form distinct clusters, implying separation between the phase-wise activations. That is, the model representations for non-similar phase-wise datasets are well-separated. The separation between model representations results in increased weight interference during Phase 2 -- leading to a decline in EA.

\paragraph{Visualizing Variance in Model Representations.} Figure~\ref{fig::tsne} provides an intuition for the correlation between phase-wise datasets and the decline in English ability. To further understand the layer-wise behavior of the hidden activations, similar to \citet{chang-etal-2022-geometry}, we compute covariance matrices $\Sigma_\Theta$ for each $X_\Theta$. Intuitively, $\Sigma_\Theta$ captures the variance in different directions for representations of hidden activations for $\Theta$.

%
\begin{figure}
    \centering
    \includegraphics[width=\columnwidth]{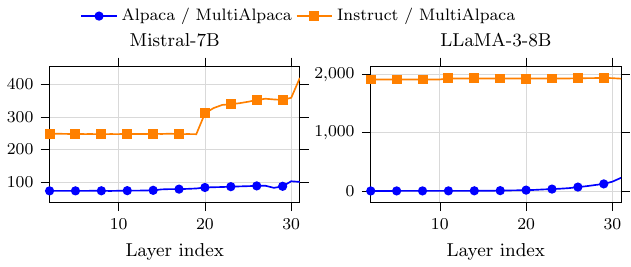}
    \caption{\textbf{Visualizing variance in model representations.} We plot $\|\Sigma_{\text{Phase 2}} - \Sigma_{\text{Phase 1}}\|_2$ across layers to measure representation shift. Dissimilar datasets (e.g., Instruct \& \malp) induce substantially larger variation than similar ones (e.g., \alp \& \malp). For \llm, this variation is consistently high across layers, whereas for \mis, it is more localized and increases primarily in higher layers.}
    \label{fig:variance_1}
\end{figure}
%

We first compute the mean centered activation matrix $\Bar{X}_\Theta = X_\Theta - \mu_\Theta$, where $\mu_{\Theta} = 1/l\sum_{i=1}^{l} X_{\Theta}^{(i)}$. Next, we derive $\Sigma_\Theta = \frac{1}{l-1} \cdot \Bar{X}_\Theta^T \Bar{X}_\Theta \in \mathbb{R}^{d\times d}$. To compare the layer-wise variance in representations, we compute the L2-Norm of the difference of the matrices $\Sigma_{\mis}$ (Figure~\ref{fig:variance_1}~\textbf{(left)}) or $\Sigma_{\llm}$ (Figure~\ref{fig:variance_1}~\textbf{(right)}) when continually fine-tuned on \alp\ \& \malp\ \textcolor{blue}{(blue lines)} or Instruct \& \malp\ \textcolor{red}{(red lines)}.

From the figures, we see clear evidence of representational change, both in terms of the magnitude of the change and the subset of layers that show a greater change. For \mis, the Phase 2 model after CFT with Instruct \& \malp, shows 3 to 4 times more variation in its representations compared to the model with \alp\ \& \malp\ phase-wise datasets. This gap is significantly larger for \llm.

%
\begin{table*}[t]
    \centering
    \begin{adjustbox}{max width=\textwidth}
    \begin{tabular}{cc|cccccc|ccccc|c}
    \toprule
          \multicolumn{2}{c|}{\textbf{CFT Setup}} & \multicolumn{6}{|c|}{\textbf{English Ability (EA)}} & \multicolumn{5}{c|}{\textbf{Multilingual Ability (MA)}} & \textbf{Combined} \\ 
           \midrule
       \multirow{2}{*}{}   & \textbf{Mitigating} & \multirow{1}{*}{\texttt{IFEval}}  & \multirow{1}{*}{\texttt{Alpaca Eval}} &  \multirow{1}{*}{\texttt{MMLU}} & \multirow{1}{*}{\texttt{HellaSwag}} &  \texttt{XLSUM\_en} & \textbf{Avg} & \multirow{1}{*}{\texttt{MLQA}} & \multirow{1}{*}{\texttt{XLSUM}} & \multirow{1}{*}{\texttt{XQUAD}} & \texttt{GMMLU} & \textbf{Avg} & {\textbf{Avg}} \\  
    &  \textbf{Strategy} & $(\uparrow)$ & $(\uparrow)$ & $(\uparrow)$ & $(\uparrow)$ & $(\uparrow)$ &  $(\uparrow)$ &  $(\uparrow)$ & $(\uparrow)$ & $(\uparrow)$ & $(\uparrow)$ &  $(\uparrow)$  & $(\uparrow)$ \\
    \midrule
 \multirow{7}{*}{\rotatebox{90}{\mis}}   &   -- & 0.462 & 0.15 & 0.533 & 0.416 & 0.10 & 0.332 & 0.307 & 0.033 & 0.436 & 0.430 &  0.302 & 0.317 \\
    &   \texttt{LF\_H1} & 0.456 & 0.03  & 0.497 &  0.598 &  0.10 & 0.336 &  0.176  & 0.016   &  0.215 & 0.428 & 0.209 & 0.273 \\  
                             &   \texttt{LF\_H2} & 0.364 & 0.12 & 0.364 &   0.504  & 0.12 & 0.294 & 0.213 & 0.014 &  0.442  & 0.384  &  0.263 &  0.279 \\  
                            &    \texttt{Spectrum} & 0.435  &  \textbf{0.24}  &   0.488  & 0.524 &  \textbf{0.13} & 0.363  & \textbf{0.317} &  \textbf{0.083} & 0.176 & 0.370 &  0.237 & 0.30 \\ 
                                 & \texttt{GR\_5} & 0.540 & 0.17 & 0.540  & 0.611   & 0.11 & 0.394  &  {0.311} & 0.008 & 0.428 & 0.445 & 0.298 & \textbf{0.346} \\  
                           &    \texttt{GR\_10} & 0.567 & 0.12 & 0.567 & 0.594 &0.12   & 0.394 &  0.213 & 0.007 & 0.427 &  \textbf{0.450} & 0.274 & 0.334  \\  
                             &    \texttt{ER\_10} & \textbf{0.593}  &  {0.08}  &   \textbf{0.580}  & \textbf{0.635} & \textbf{0.13} & \textbf{0.404}  & 0.249 &  0.008 & 0.398 & 0.448  & 0.276& 0.340\\ 
                             &    \texttt{LoRA} & 0.383  &  0.09  &    0.579  & 0.625 & 0.03 & 0.341 & 0.289  &  {0.043} & \textbf{0.518} & 0.435 & \textbf{0.321} & 0.331\\
    \midrule
  \multirow{7}{*}{\rotatebox{90}{\llm}}   &   -- & 0.182 & 0.10 & 0.239 & 0.278  &  0.09 & 0.178 & 0.321 & 0.030 &  0.417  & 0.440 & 0.302 & 0.240 \\
 &    \texttt{LF\_H1} & 0.303  & 0.0  & 0.231 & 0.275 &  0.01   & 0.164  & 0.368  & 0.037 & 0.505 & 0.237 & 0.287 & 0.225 \\  
                        &  \texttt{LF\_H2}  &  0.380 &  0.06  & 0.485 & 0.525 & 0.08 &  0.306 &  0.400 & 0.038 & 0.505 & 0.338 &0.320 & 0.313\\  
                        &   \texttt{Spectrum} &  0.409 &  0.09  &  0.612   & 0.524 &  0.01 & 0.329    & 0.429 & \textbf{0.056}  & 0.086 &\textbf{0.472} & 0.261 & 0.295 \\ 
                        &  \texttt{GR\_5} &  0.269 & 0.01  & 0.516 & 0.316 & 0.07 & 0.236 & \textbf{0.437}  & 0.019   & \textbf{0.593} & 0.342 & 0.348  & 0.292\\  
                        &  \texttt{GR\_10}  & 0.264  & \textbf{0.12} & 0.229 & 0.250  & 0.0 &  0.173 &  0.254           &  0.009   & 0.314 &  0.238 & 0.204 & 0.189 \\ 
                        &   \texttt{ER\_10} & \textbf{0.420} &  {0.02}  &   \textbf{0.603}   &  \textbf{0.561} & \textbf{0.12} &  \textbf{0.345} & 0.434  & 0.025 &   0.53 & 0.448 & \textbf{0.359} & \textbf{0.352} \\
                        &   \texttt{LoRA} &  0.196 &  0.0  &   0.280   & 0.235 & 0.0  &  0.142 &  0.007 & 0.008  & 0.005 & 0.278 & 0.075 & 0.109  \\
    \bottomrule
    \end{tabular}
    \end{adjustbox}
    \caption{\textbf{English Ability (EA) and Multilingual Ability (MA) results for our mitigating strategies.} These comprise Generative Replay (\texttt{GR\_5} \& \texttt{GR\_10}), English Replay (\texttt{ER\_10}) and Layer Freezing (\texttt{LF\_H1},  \texttt{LF\_H2} \& \texttt{Spectrum}). We  use \texttt{LoRA}~\cite{hu2022lora} as a baseline strategy. For \texttt{ER\_10}, we use the English dataset used in \texttt{GR} with original responses. \textit{The Phase 1 dataset is Instruct for each row, while Phase 2 is \malp.} The first row for both \mis\ and \llm\ provides numbers for Instruct-\malp\ (from Table~\ref{tab:results_main} \& ~\ref{tab:results_main_MA}).}
    \label{tab:results_interventions}
\end{table*}

\label{sec:eval_langadapt}

\section{Mitigating Strategies for CFT}\label{sec:mitigation}

To mitigate EA decline, we explore two tailored CFT techniques\footnotemark: Distribution Replay and Layer Freezing. In Distribution Replay, we study Generative Replay (\texttt{GR}), a new English data generation method inspired by dataset similarity and English ability (\S\ref{subsec::ta_la_results}), and English Replay (\texttt{ER}), which replays parallel English data of Phase 2's distribution. In Layer Freezing (\texttt{LF}), we identify layers to freeze during Phase 2 fine-tuning using specific heuristics.

\footnotetext{In \S\ref{sec:cft_comparison}, we discuss how MAD-X~\cite{pfeiffer2020madxadapterbasedframeworkmultitask} and other PEFT-based methods~\cite{badola-etal-2023-parameter} are not suitable for our setting.}

\subsection{Distribution Replay}
Typically, Generative Replay (\texttt{GR}) is a technique that generates data from past distributions to be used alongside new task data for the continual fine-tuning of a model on a new task \cite{shin2017continuallearningdeepgenerative}. However, from \S\ref{subsec::ta_la_results}, we do not see a decline in English ability if the phase-wise datasets encode similar instructions. Based on this, we use the Phase 1 model to generate responses, in English, from the English counterpart of the multilingual dataset used for fine-tuning in Phase 2. The intuition is that the generated dataset may bridge the distributions of Phase 1 and Phase 2.  

During Phase 2 fine-tuning, we include varying quantities of this generated data: specifically, 5\% (\texttt{GR\_5}) and 10\% (\texttt{GR\_10}), of the Phase 2 dataset. We also fine-tune the models with a similar sized subset of the English counterpart with original responses\footnote{This dataset may not be available for all multilingual datasets, such as Aya~\cite{singh2024ayadatasetopenaccesscollection}. While instructions can be translated into English, translating responses is often impractical. Thus, \texttt{ER} is the best-case scenario for \texttt{GR}.}. We refer to this mitigating strategy as English Replay (\texttt{ER\_10}).

\subsection{Layer Freezing}
Model regularization is an effective technique to mitigate the drop in the previous task's performance in continual learning~(e.g., EWC~\cite{kirkpatrick2017overcoming}). However, this is computationally inefficient as it requires using both the old and new sets of parameters. Instead, we use Layer Freezing (\texttt{LF}), a relatively efficient technique for use as a `regularizer' to preserve English ability during Phase 2. We consider the following variations to select the set of layers to freeze. These variants allow us to study how different criteria for selecting frozen layers affect the trade-off between preserving English ability and enabling multilingual adaptation.
\begin{enumerate}
    \item \texttt{LF\_H1}: freezing a random set of 10 layers of the model from Phase 1 to be fine-tuned in Phase 2.
    \item \texttt{LF\_H2}: freezing the top-10 layers that have changed the most during Phase 1 fine-tuning (e.g., \mis\ Base to \misi). We select layers separately for Key, Query, and Value, for each attention head.
    \item \texttt{Spectrum}~\cite{hartford2024spectrum}: freeze the "most informative" layers of the Phase 1 model based on their signal-to-noise ratio (refer to \S\ref{app::spectrum}). 
\end{enumerate}

 We present our results in Table \ref{tab:results_interventions} for both \texttt{GR} and \texttt{LF}.  We define a \textbf{baseline} in which we use \texttt{LoRA}~\cite{hu2022lora}\footnotemark for continually fine-tuning in Phase 2. We perform \texttt{LoRA} fine-tuning with rank \texttt{64} and quantisation \texttt{bfloat16}.

\footnotetext{Parameter efficient techniques like LoRA~\cite{hu2022lora} are also widely used to efficiently fine-tune LLMs on multilingual data. However, such techniques also show \emph{forgetting} on English \cite{aggarwal2024maple} after Phase 2.} 

\subsection{Results Discussion\label{subsec::main_MS_discussion}}
From Table~\ref{tab:results_interventions}, we see that \texttt{GR}, \texttt{ER}, and \texttt{LF} mitigate the decline in EA and also show gains in MA.

\paragraph{Distribution Replay.} \texttt{ER\_10} demonstrates the best performance in both English and combined ability, with EA scores of 0.404 for \mis\ and 0.345 for \llm, and the best combined average. \texttt{GR\_5} also excels in multilingual tasks, performing similar to \texttt{ER\_10}: 0.298 vs. 0.276 for \mis\ and 0.348 vs. 0.359 for \llm. 
\texttt{GR\_5} also performs reasonably well on English tasks, achieving scores of 0.394 and 0.236 for \mis\ and \llm, respectively, making it a competitive strategy.

\paragraph{Layer Freezing.} Compared to \texttt{ER} and \texttt{GR}, \texttt{LF\_H1}, \texttt{LF\_H2}, and \texttt{Spectrum} show mixed results. \texttt{LF\_H2} performs better than \texttt{LF\_H1}. \texttt{Spectrum}'s EA scores are better than \texttt{LF\_H1} and \texttt{LF\_H2}, but suffers from lower multilingual numbers.

\paragraph{Additional Discussion \& Results.} In \S\ref{app::mitigation_ablations}, we also present EA and MA results for \mis\ Instruct-\morca\ for our mitigating strategies. Here, \texttt{LF}, particularly \texttt{Spectrum}, performs better than the other strategies. Furthermore, in \S\ref{app::compute_analysis}, we analyze the computational cost of these strategies over the baseline CFT setup.

\subsection{Controlling Representation Drift under Mitigation}

\begin{figure}
    \centering
        \includegraphics[width=\columnwidth]{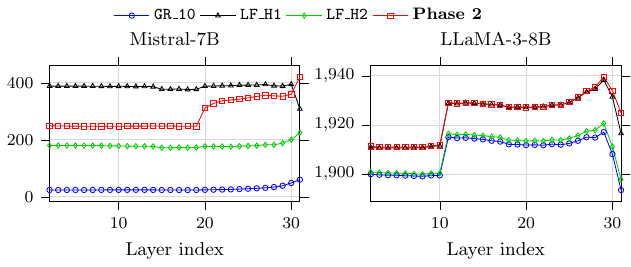}
        \caption{\textbf{Layer-wise covariance shift under mitigation strategies in Phase~2 CFT.} We plot $\|\Sigma_{\text{Phase 2}} - \Sigma_{\text{Phase 1}}\|_2$ across layers. The vanilla Phase~2 model (Figure~\ref{fig:variance_1}) exhibits a sharp increase in drift in higher layers. Replay (\texttt{GR}) substantially reduces the magnitude of this drift across all layers, while freezing (\texttt{LF\_H2}) confines it to a smaller subset of layers.}
    \label{fig:variance_MS}
\end{figure}

Building on Figure~\ref{fig:variance_1}, we examine how mitigation strategies affect representation drift relative to the vanilla Phase~2 baseline (\mis\ Instruct $\rightarrow$ \malp). As in \S\ref{subsec::vs_decline}, we measure drift as $\|\Sigma_{\text{Phase 2}} - \Sigma_{\text{Phase 1}}\|_2$, where larger values indicate greater deviation from Phase~1 representations.

Consistent with earlier observations, the mitigation strategies explicitly designed to curb representational change (\texttt{LF} \& \texttt{GR}) exhibit lower drift than the baseline model across layers. Replay-based methods (\texttt{GR}) reduce the overall magnitude of drift while preserving its layer-wise structure, whereas freezing-based methods (\texttt{LF\_H2}) constrain the drift to a subset of layers.

Notably, generative replay remains closest to the Phase~1 representation geometry among all strategies. This closer alignment correlates with improved task and language performance compared to the vanilla Phase~2 model (Table~\ref{tab:results_main} and Table~\ref{tab:results_main_MA}). These results reinforce our central hypothesis: in two-phase CFT with dissimilar datasets, representation shift is the primary driver of English degradation, and explicitly controlling this shift preserves alignment with Phase~1 representations, leading to improved English retention without sacrificing multilingual gains.

\section{Conclusion \& Future Work}
We study the role of dataset similarity in a novel two-phase continual fine-tuning (CFT) for multilingual adaptation. Across \mis\ and \llm, we show that alignment between Phase~1 and Phase~2 datasets is critical: similar datasets preserve English ability while improving multilingual performance, whereas misaligned datasets lead to degradation. We identify representation shift as the underlying mechanism driving this behavior. In particular, misaligned fine-tuning induces large covariance shifts in hidden representations, concentrated in higher layers. We show that mitigation strategies such as generative replay and layer freezing improve English retention by controlling this drift, either by globally reducing its magnitude or by constraining it to specific layers, while preserving multilingual gains.

\paragraph{Future Work.} Our results suggest several directions for future research. First, designing adaptive mitigation strategies that dynamically control representation drift during training could provide stronger and more efficient trade-offs than fixed heuristics~\cite{zhang2026dynamic}. Second, developing principled measures of instruction-level dataset similarity may enable better prediction and prevention of degradation. Finally, exploring parameter-efficient approaches that explicitly regularize representation geometry could reduce the computational overhead of current methods while maintaining their effectiveness.

\section{Limitations}
Our study has several limitations. First, we rely on DES and MPD as proxies for dataset similarity; while effective in our setting, these metrics may not capture all nuances of instruction-level similarity. Second, our experiments are limited to \mis\ and \llm, and the observed trends may not fully generalize to models with different architectures, scales, or training paradigms. Similarly, the effectiveness of our mitigation strategies (i.e., replay and layer freezing) may vary across datasets and model families. Third, our experimental setup is constrained by computational resources. We were unable to explore larger models, longer training schedules, or broader hyperparameter sweeps, which may further influence the observed trade-offs. Fourth, some mitigation strategies impose additional requirements. In particular, the best-performing method, \texttt{ER\_10}, assumes access to parallel data, which may not always be available in practice. Finally, our evaluation focuses on specific benchmarks for task and language ability. While these provide controlled comparisons, they may not fully capture real-world performance across diverse applications.

\newpage
\bibliography{ref}

\newpage
\appendix

\newpage
\appendix

\section{Training Details \label{app::training}}

\subsection{Hyperparameters for Fine-tuning and Training Setup}

\begin{table}[h]
    \centering
    \renewcommand{\thetable}{\thesection\arabic{table}}\setcounter{table}{0}
    \begin{adjustbox}{max width=\columnwidth}
    \begin{tabular}{ll}
    \toprule
     \textbf{Hyperparameter}    &  \textbf{Value}\\
     \midrule
        Learning Rate &  $1\times 10^{-6}$  \\
        Epochs    &    4   \\
        Global Batch size  &  16 \\
        Scheduler  & Cosine  \\
        Warmup     &   Linear  \\
        Warmup Steps & 10   \\
        Optimizer  &  AdamW~\cite{loshchilov2019decoupled}  \\
        Weight Decay &   0  \\
    \bottomrule
    \end{tabular}
    \end{adjustbox}
    \smallskip
    \caption{Hyperparameters for our Two-phase Continual Fine-tuning (CFT)}
    \label{tab:hyperparameters}
\end{table}

\subsection{Fine-tuning Datasets\label{app::fine-tuning-datasets}}
\orca\ is an English-only self instruct dataset, created to best mimic the \textsc{Orca} dataset~\cite{mukherjee2023orca}, which is not publicly available. To create the multilingual version of \orca, namely \morca, we follow \citet{sphinx} to generate selective translations for a subset of \orca. The subset contains 50k samples from the \orca\ dataset and we selectively translate them to 11 languages which are also in \malp. In total, we generate 550k examples for all languages. 

\subsection{Evaluation Tasks}\label{app: eval details} In this paper, we consider two sets of benchmarks to evaluate task and language ability. We explain them briefly next.


\paragraph{English Ability (EA).}  
To quantify an LLM's task ability, we evaluate Phase 1 and Phase 2 models on the following tasks:

\begin{enumerate}[leftmargin=*,noitemsep]
    \item \texttt{IFEval}~\cite{zhou2023instruction}: Instruction-Following Evaluation (\texttt{IFEval}) asses the ability of an LLM to follow natural language instructions. It comprises 500 verifiable instructions (e.g., ``\emph{mention the keyword AI 3 times}''). 
    We choose \texttt{IFEval} as the instructions are verifiable and also test an LLM's context understanding. 

    \item \texttt{Alpaca Eval}~\cite{li2023alpacaeval}: This is an LLM-based automatic evaluator for instruction following models, to measure task ability. Like \citet{aggarwal2024maple}, we evaluate our CFT models against \emph{text-davinci-003} responses on 800 instructions and use GPT4 (\emph{gpt-4-32k)} as the evaluator.

    \item \texttt{\texttt{MMLU}}~\cite{hendryckstest2021}: Massive Multitask Language Understanding (\texttt{\texttt{MMLU}}) is a benchmark to assess an LLM's knowledge and problem-solving abilities. It includes 57 subjects across domains like STEM, or law, with 16k MCQs in total. 

    \item \texttt{\texttt{HellaSwag}}~\cite{zellers2019hellaswag}: This is a popular benchmark to evaluate the commonsense reasoning ability of an LLM. \texttt{\texttt{HellaSwag}}'s test split contains 10k samples in total. 
\end{enumerate}

\paragraph{Multilingual Ability (MA).} 
To quantify an LLM's language ability, we evaluate our fine-tuned models on three benchmark datasets comprising two multilingual generative tasks: question answering and summarization.

\begin{itemize}[leftmargin=*,noitemsep]
    \item \textbf{Question Answering:} \texttt{\texttt{MLQA}} \cite{lewis2019mlqa} contains 5k extractive question-answering instances in 7 languages. The \texttt{\texttt{XQuAD}} dataset \cite{Artetxe:etal:2019} consists of a subset of 240 paragraphs and 1190 question-answer pairs across 11 languages. 
    \item \textbf{Summarisation:} \texttt{XLSUM}  \cite{hasan-etal-2021-xl} spans 45 languages, and we evaluate our models in Arabic, Chinese-Simplified, English, French, Hindi, Japanese, and Spanish. 
\end{itemize}

\section{Evaluating Multilingual Ability for Continual Fine-tuning\label{app::more_Evals} }

\begin{table*}[t!]
    \centering
        \centering\renewcommand{\thetable}{\thesection\arabic{table}}
    \begin{adjustbox}{max width=\textwidth}
    \begin{tabular}{ccccccccccccccc}
    \toprule
    \multicolumn{15}{c}{\textbf{Two-phase Continual Fine-tuning}} \\
    \midrule
    \multirow{2}{*}{\textbf{Model}}     & \textbf{Phase 1 (P1)} & \textbf{Phase 2 (P2)} & \multicolumn{2}{c}{\texttt{IFEval} $(\uparrow)$}  & \multicolumn{2}{c}{\texttt{Alpaca Eval} $(\uparrow)$} &  \multicolumn{2}{c}{\texttt{MMLU} $(\uparrow)$} & \multicolumn{2}{c}{\texttt{HellaSwag} $(\uparrow)$} & \multicolumn{2}{c}{\texttt{XLSUM\_en} $(\uparrow)$} & \multicolumn{2}{c}{\textbf{Average}} \\
    & \textbf{Dataset} & \textbf{Dataset} & \textbf{P1} & \textbf{P2}  & \textbf{P1} &  \textbf{P2} & \textbf{P1} & \textbf{P2}  & \textbf{P1} & \textbf{P2} & \textbf{P1} & \textbf{P2} & \textbf{P1} & \textbf{P2} \\ 
    \midrule
    \multirow{2}{*}{\mis} & \orca & \multirow{1}{*}{\morca} & 0.494  &  \textcolor{pastelred}{0.482}  & 0.31  &  \textcolor{green!80!black}{0.32}  &  {0.601}  & \textcolor{pastelred}{0.582} &      0.612 &  \textcolor{pastelred}{0.562 }  & 0.02 & \textcolor{green!80!black}{0.12} & 0.407 & \textcolor{green!80!black}{0.413 }\\  
    & Instruct & \morca  & 0.550  & \textcolor{pastelred}{0.426} & {0.35} &  \textcolor{pastelred}{0.06} & 0.575    & \textcolor{pastelred}{0.507} & {0.641} &   \textcolor{pastelred}{0.509} & 0.13 &  \textcolor{pastelred}{0.0} & 0.450  & \textcolor{pastelred}{0.30} \\  
    \midrule
    \multirow{2}{*}{\llm} & \orca & \morca  &  0.377 & \textcolor{green!80!black}{0.425}  & 0.09  & \textcolor{pastelred}{0.07} &  0.579 &  \textcolor{green!80!black}{0.599}  & 0.571 &  \textcolor{pastelred}{0.564} & 0.07 & \textcolor{green!80!black}{0.12}  &  0.337 &   \textcolor{green!80!black}{0.356} \\ 
    & Instruct & \morca & {0.735}   & \textcolor{pastelred}{0.205}  & 0.14 & \textcolor{pastelred}{0.0} & 0.340 &  \textcolor{pastelred}{0.236}  & 0.533  &  \textcolor{pastelred}{0.250} & 0.11 & \textcolor{pastelred}{0.0} & 0.372 & \textcolor{pastelred}{0.138} \\
    \bottomrule
    \multicolumn{15}{c}{\textbf{Dataset Mixture}} \\
    \midrule
    \textbf{Model} & \multicolumn{2}{c}{\textbf{Dataset Mixture}} & \multicolumn{2}{c}{\texttt{IFEval} $(\uparrow)$} & \multicolumn{2}{c}{\texttt{Alpaca Eval} $(\uparrow)$} & \multicolumn{2}{c}{\texttt{MMLU} $(\uparrow)$} & \multicolumn{2}{c}{\texttt{HellaSwag} $(\uparrow)$} & \multicolumn{2}{c}{\texttt{XLSUM\_en} $(\uparrow)$} & \multicolumn{2}{c}{\textbf{Average}} \\
    \midrule
    \mis & \multicolumn{2}{c}{\orca\ \morca} & \multicolumn{2}{c}{0.228} & \multicolumn{2}{c}{0.035} & \multicolumn{2}{c}{0.284}  & \multicolumn{2}{c}{0.444}  & \multicolumn{2}{c}{0.02} & \multicolumn{2}{c}{0.202} \\ 
    \llm & \multicolumn{2}{c}{\orca\ \morca} & \multicolumn{2}{c}{0.248} & \multicolumn{2}{c}{0.072} & \multicolumn{2}{c}{0.484}  & \multicolumn{2}{c}{0.473}  & \multicolumn{2}{c}{0.0} &  \multicolumn{2}{c}{0.255} \\
    \bottomrule
    \end{tabular}
    \end{adjustbox}
    \caption{English Ability results for two-phase Continual Fine-tuning (CFT). With \textcolor{green!80!black}{green}, we highlight an increase in a model's task ability post P2 fine-tuning. Likewise, \textcolor{pastelred}{red} highlights a decline in a model's task ability. }
    \label{tab:results_main_ablation}
\end{table*}

%
\begin{table*}[t!]
    \centering\renewcommand{\thetable}{\thesection\arabic{table}}
    \begin{adjustbox}{max width=\textwidth}
    \begin{tabular}{ccccccccccccc}
    \toprule
    \multicolumn{13}{c}{\textbf{Two-phase Continual Fine-tuning}} \\
    \toprule
    \multirow{2}{*}{\textbf{Model}}     & \textbf{Phase 1 (P1)} & \textbf{Phase 2 (P2)} & \multicolumn{2}{c}{\texttt{MLQA} $(\uparrow)$}  & \multicolumn{2}{c}{\texttt{XLSUM} $(\uparrow)$} &  \multicolumn{2}{c}{\texttt{XQuAD} $(\uparrow)$} & \multicolumn{2}{c}{\texttt{GMMLU} $(\uparrow)$} & \multicolumn{2}{c}{\textbf{Average}} \\
    & \textbf{Dataset} & \textbf{Dataset} & \textbf{P1} & \textbf{P2}  & \textbf{P1} &  \textbf{P2} & \textbf{P1} & \textbf{P2} & \textbf{P1} & \textbf{P2}  & \textbf{P1} & \textbf{P2}   \\
      \midrule
    \multirow{2}{*}{\mis} & \orca & \multirow{1}{*}{\morca} & 0.435  &  \textcolor{pastelred}{0.360} &  0.007 &  \textcolor{green!80!black}{0.008}  &  0.556  & \textcolor{green!80!black}{0.643} & 0.433 & \textcolor{pastelred}{0.312} & 0.358 & \textcolor{pastelred}{0.331} \\
    & Instruct & \morca & 0.246  & \textcolor{pastelred}{0.155} & 0.012  & \textcolor{green!80!black}{0.040} &  0.351  & \textcolor{pastelred}{0.323} & 0.440 & \textcolor{pastelred}{0.279} & 0.262 & \textcolor{pastelred}{0.20} \\  
    \midrule
    \multirow{2}{*}{\llm} & \orca & \morca & 0.401 & \textcolor{green!80!black}{0.453} & 0.017 &   \textcolor{pastelred}{0.006} &  0.499 &    \textcolor{green!80!black}{0.531} & 0.242 & \textcolor{green!80!black}{0.513} & 0.290 & \textcolor{green!80!black}{0.376} \\  
    & Instruct & \morca & {0.609}  & \textcolor{pastelred}{0.604} & {0.048} & \textcolor{green!80!black}{ 0.048} & {0.712} & \textcolor{green!80!black}{0.713} & 0.250 & \textcolor{pastelred}{0.233} &0.405 & \textcolor{pastelred}{0.40}
                              \\ 
    \bottomrule
    \multicolumn{13}{c}{\textbf{Dataset Mixture}} \\
    \midrule
    \textbf{Model} & \multicolumn{2}{c}{\textbf{Dataset Mixture}} & \multicolumn{2}{c}{\texttt{MLQA} $(\uparrow)$} & \multicolumn{2}{c}{\texttt{XLSUM} $(\uparrow)$} & \multicolumn{2}{c}{\texttt{XQuAD} $(\uparrow)$} & \multicolumn{2}{c}{\texttt{GMMLU} $(\uparrow)$}  & \multicolumn{2}{c}{\textbf{Average}} \\
    \midrule
    \mis & \multicolumn{2}{c}{\orca\ \morca} & \multicolumn{2}{c}{0.201} & \multicolumn{2}{c}{0.128} & \multicolumn{2}{c}{0.071}  & \multicolumn{2}{c}{0.277} & \multicolumn{2}{c}{0.169} \\ 
    \llm & \multicolumn{2}{c}{\orca\ \morca} & \multicolumn{2}{c}{0.224} & \multicolumn{2}{c}{0.034} & \multicolumn{2}{c}{0.091}  & \multicolumn{2}{c}{0.364} & \multicolumn{2}{c}{0.178}\\
    \bottomrule
    \end{tabular}
    \end{adjustbox}
    \caption{Multilingual Ability results for two-phase Continual Fine-tuning (CFT). With \textcolor{green!80!black}{green}, we highlight an increase in a model's Multilingual ability post Phase 2 fine-tuning. Likewise, \textcolor{pastelred}{red} highlights a decline in a model's Multilingual ability. }
    \label{tab:results_main_LA_supp}
\end{table*}


\paragraph{English Ability.} Table~\ref{tab:results_main_ablation} present the english ability numbers of our ablations on the \orca-\morca and Instruct-\morca datasets using \mis\ and \llm\ models. When the datasets are pairwise not similar, i.e., Instruct-\morca, \mis\ shows a significant decline in the \textit{average} English ability, from 0.450 in Phase 1 to 0.30 in Phase 2. Likewise, \llm\ also experiences a decrease, dropping from 0.372 to 0.138 on average.

In contrast, when the pairwise datasets are similar, i.e., \orca\ and \morca, \mis\ sees a marginal increase between the phases $(0.407 \rightarrow 0.413)$, on average. \llm's performance sees an improvement in the average English ability, from 0.337 to 0.356.

\paragraph{Multilingual Ability.} Table~\ref{tab:results_main_LA_supp} tabulates the results for multilingual ability. We see an improvement in the \emph{average} multilingual ability for the \orca-\morca\ dataset pair for \llm. For \mis, there is a marginal drop $(0.358 \to 0.331)$. For Instruct-\morca, with \llm, the average multilingual ability is virtually the same across tasks (0.405 vs. 0.40). However, for \mis, we see a slight drop in the average language ability, driven primarily due to a decline in performance for \texttt{MLQA}.


%
\begin{table*}[t!]
    \centering\renewcommand{\thetable}{\thesection\arabic{table}}
    \begin{adjustbox}{max width=\textwidth}
    \begin{tabular}{ccccccccccccc}
    \toprule
    \multirow{2}{*}{\textbf{Model}}     & \textbf{Phase 1 (P1)} & \textbf{Phase 2 (P2)} & \multicolumn{2}{c}{\texttt{IFEval} $(\uparrow)$}  & \multicolumn{2}{c}{\texttt{Alpaca Eval} $(\uparrow)$} &  \multicolumn{2}{c}{\texttt{MMLU} $(\uparrow)$} & \multicolumn{2}{c}{\texttt{HellaSwag} $(\uparrow)$} & \multicolumn{2}{c}{\textbf{Average}} \\
    & \textbf{Dataset} & \textbf{Dataset} & \textbf{P1} & \textbf{P2}  & \textbf{P1} &  \textbf{P2} & \textbf{P1} & \textbf{P2}  & \textbf{P1} & \textbf{P2}  & \textbf{P1} & \textbf{P2} \\ 
    \midrule 
   \mis &  \malp  & \alp  &  0.245 &   \textcolor{green!80!black}{0.290} & 0.120    &  \textcolor{pastelred}{0.114} &   0.528   &  \textcolor{pastelred}{0.430} &  0.476      & \textcolor{green!80!black}{0.510}  & 0.342 &  \textcolor{pastelred}{0.336}\\ 
    \llm &  \malp  & \alp  & 0.245 &  \textcolor{green!80!black}{0.340}  &  0.038   & \textcolor{green!80!black}{0.065}  &    0.570  & \textcolor{pastelred}{0.540}  &   0.577    &  \textcolor{green!80!black}{0.590} &  0.357 &  \textcolor{green!80!black}{0.384}\\ 
     \mis   &  \morca & \orca  &  0.190 & \textcolor{green!80!black}{0.310} & 0.091 &  \textcolor{pastelred}{0.055} & 0.410 &  \textcolor{green!80!black}{0.490}  & 0.520 &  0.510  & 0.303 & \textcolor{green!80!black}{0.341} \\
        \llm    &  \morca & \orca  &  0.314 & \textcolor{green!80!black}{0.340} & 0.0 & 0.0  & 0.530  &  \textcolor{green!80!black}{0.540}  & 0.522 &  \textcolor{green!80!black}{0.590}  & 0.342 & \textcolor{green!80!black}{0.368}\\
    \bottomrule
    \end{tabular}
    \end{adjustbox}
    \caption{English Ability results for two-phase Continual Fine-tuning (CFT)  }
    \label{tab:results_flip_TA}
\end{table*}

%
\begin{table*}[t!]
    \centering\renewcommand{\thetable}{\thesection\arabic{table}}
    \begin{adjustbox}{max width=\textwidth}
    \begin{tabular}{ccccccccccc}
    \toprule
    \multirow{2}{*}{\textbf{Model}}     & \textbf{Phase 1} & \textbf{Phase 2} & \multicolumn{2}{c}{\texttt{MLQA} $(\uparrow)$}  & \multicolumn{2}{c}{\texttt{XLSUM} $(\uparrow)$} &  \multicolumn{2}{c}{\texttt{XQuAD} $(\uparrow)$} & \multicolumn{2}{c}{\textbf{Average}} \\
    & \textbf{Dataset} & \textbf{Dataset} & \textbf{Phase 1} & \textbf{Phase 2}  & \textbf{Phase 1} &  \textbf{Phase 2} & \textbf{Phase 1} & \textbf{Phase 2} & \textbf{Phase 1} & \textbf{Phase 2}   \\
    \midrule 
    \mis &  \malp  & \alp  & 0.122  &   \textcolor{green!80!black}{0.230} &  0.021   & \textcolor{green!80!black}{0.030 }  &   0.122  &  0.090 &   0.088    &  \textcolor{green!80!black}{0.116}   \\  
   \llm &  \malp  & \alp  & 0.363  &  \textcolor{pastelred}{0.340}  &   0.048  &  \textcolor{green!80!black}{0.040} &    0.058  &  \textcolor{pastelred}{0.030} &   0.157    &   \textcolor{pastelred}{0.134}  \\ 
    \mis   &  \morca & \orca  &  0.165 &  \textcolor{pastelred}{0.160} & 0.077 & \textcolor{green!80!black}{0.070} &  0.140 &  \textcolor{green!80!black}{0.180}  & 0.127 & \textcolor{green!80!black}{0.137} \\
     \llm   &  \morca & \orca  &  0.057  & 0.0 & 0.038 & 0.0 &  0.047 & 0.0  &  0.047   & 0.0  \\
   \bottomrule
    \end{tabular}
    \end{adjustbox}
    \caption{Multilingual Ability results for two-phase Continual Fine-tuning (CFT)}
    \label{tab:results_flip_LA}
\end{table*}

Furthermore, Table~\ref{tab:results_mlqa}, Table~\ref{tab:results_xlsum}, and Table~\ref{tab:results_xquad} present the language-specific results for MLQA, XLSUM, and XQuAD, respectively. 


%
\begin{table*}[t!]
    \centering\renewcommand{\thetable}{\thesection\arabic{table}}
    \begin{adjustbox}{max width=\textwidth}
    \begin{tabular}{ccc|cccccc|cccccc}
    \toprule
    \multirow{2}{*}{\textbf{Model}}     & \textbf{Phase 1} & \textbf{Phase 2} & \multicolumn{12}{c}{\texttt{MLQA}}    \\
    & \textbf{Dataset} & \textbf{Dataset} & \multicolumn{6}{c|}{\textbf{Phase 1}} & 
  \multicolumn{6}{c}{\textbf{Phase 2}}   \\
  &  &    &  \textbf{ar}  & \textbf{de} & \textbf{es} & \textbf{hi} & \textbf{vi} &\textbf{zh}   &  \textbf{ar}  & \textbf{de} & \textbf{es} & \textbf{hi} & \textbf{vi} & \textbf{zh} \\
    \midrule
    \multirow{2}{*}{\mis} & \alp & \multirow{4}{*}{\malp} & 0.143 & 0.337 & 0.331 & 0.149 & 0.385 & 0.031  &  0.172 & 0.485 & 0.529 & 0.196 & 0.336 & 0.009 \\  
                             & Instruct &  & 0.113 & 0.440 & 0.395 & 0.088 & 0.369 & 0.073 & 0.228 & 0.456 & 0.529 & 0.279 & 0.327 & 0.0222   \\  
    \multirow{2}{*}{\llm} & \alp & & 0.320 & 0.538 & 0.563 & 0.438 & 0.611 & 0.155 &    0.552 & 0.672 & 0.765 & 0.573 & 0.784 & 0.237 \\  
                              & Instruct &  & 0.549 & 0.701 & 0.769 & 0.624 & 0.788 & 0.192 & 0.316 & 0.453 & 0.526 & 0.137 & 0.464 & 0.028   \\  
    \midrule
    \multirow{2}{*}{\mis} & \orca & \multirow{4}{*}{\morca} & 0.374 & 0.504 & 0.511 & 0.395 & 0.600 & 0.226 &0.298& 0.506& 0.572& 0.274& 0.481& 0.030  \\  
                             & Instruct &  & 0.113 & 0.440 & 0.395 & 0.088 & 0.369 & 0.073 & 0.115& 0.253& 0.213& 0.088& 0.222& 0.038 \\  
    \multirow{2}{*}{\llm} & \orca &  & 0.262 & 0.545 & 0.565 & 0.369 & 0.568 & 0.099 &   0.437 & 0.549 & 0.622 & 0.462 & 0.625 & 0.024 \\  
                              & Instruct &  & 0.320 & 0.538 & 0.563 & 0.438 & 0.611 & 0.155 &  0.554 & 0.701 & 0.771 & 0.625 & 0.787 & 0.188 \\  
    \bottomrule
    \end{tabular}
    \end{adjustbox}
    \caption{\texttt{MLQA}: Language Ability results for two-phase Continual Fine-tuning (CFT).}
    \label{tab:results_mlqa}
\end{table*}

%
\begin{sidewaystable*}[t!]
    \centering\renewcommand{\thetable}{\thesection\arabic{table}}
    \begin{adjustbox}{max width=\textwidth}
    \begin{tabular}{ccc|cccccc|cccccc}
    \toprule
    \multirow{2}{*}{\textbf{Model}}     & \textbf{Phase 1} & \textbf{Phase 2} & \multicolumn{12}{c}{\texttt{XLSUM}}    \\
    & \textbf{Dataset} & \textbf{Dataset} & \multicolumn{6}{c}{\textbf{Phase 1}} & 
  \multicolumn{6}{c}{\textbf{Phase 2}}   \\
  &  &    &  \textbf{Arabic}  & \textbf{Chinese\_simplified} & \textbf{french} & \textbf{Hindi} & \textbf{Japanese} &\textbf{Spanish}   & \textbf{Arabic}  & \textbf{Chinese\_simplified} & \textbf{french} & \textbf{Hindi} & \textbf{Japanese} &\textbf{Spanish} \\
    \midrule
    \multirow{2}{*}{\mis} & \alp & \multirow{4}{*}{\malp} & 0.001 & 0.012 & 0.025 & 0.001 & 0.012 & 0.023 & 0.022 & 0.034 & 0.112 & 0.016 & 0.067 & 0.106 \\  
                             & Instruct &  & 0.001 & 0.005 & 0.028 & 0.001 & 0.009 & 0.025 &    0.016 & 0.015 & 0.060 & 0.010 & 0.040 & 0.056 \\  
    \multirow{2}{*}{\llm} & \alp & &  0.005 & 0.015 & 0.071 & 0.003 & 0.037 & 0.067 &  0.003 & 0.018 & 0.073 & 0.002 & 0.041 & 0.070 \\  
                              & Instruct &  & 0.008 & 0.015 & 0.092 & 0.004 & 0.080 & 0.087  &  0.002 & 0.013 & 0.055 & 0.001 & 0.055 & 0.051   \\  
    \midrule
    \multirow{2}{*}{\mis} & \orca & \multirow{4}{*}{\morca} & 0.001 & 0.010 & 0.014 & 0.001 & 0.007 & 0.009 & 0.001 & 0.006 & 0.018 & 0.001 & 0.008 & 0.016  \\  
                             & Instruct &  &  0.001 & 0.005 & 0.028 & 0.001 & 0.009 & 0.025 &  0.007 & 0.017 & 0.092 & 0.005 & 0.030 & 0.088 \\  
    \multirow{2}{*}{\llm} & \orca &  & 0.000 & 0.003 & 0.061 & 0.000 & 0.004 & 0.035 &   0.000 & 0.003 & 0.016 & 0.001 & 0.000 & 0.013  \\  
                              & Instruct &  & 0.008 & 0.015 & 0.092 & 0.004 & 0.080 & 0.087  & 0.007 & 0.015 & 0.091 & 0.004 & 0.082 & 0.087 \\  
    \bottomrule
    \end{tabular}
    \end{adjustbox}
    \caption{\texttt{XLSUM}: Language Ability results for two-phase Continual Fine-tuning (CFT).}
    \label{tab:results_xlsum}
\end{sidewaystable*}

\begin{sidewaystable*}
        \centering\renewcommand{\thetable}{\thesection\arabic{table}}
    \begin{adjustbox}{max width=\textwidth}
    \begin{tabular}{ccc|ccccccccccc|ccccccccccc}
    \toprule
    \multirow{2}{*}{\textbf{Model}}     & \textbf{Phase 1} & \textbf{Phase 2} & \multicolumn{22}{c}{\texttt{XQuAD}}    \\
    & \textbf{Dataset} & \textbf{Dataset} & \multicolumn{11}{c}{\textbf{Phase 1}} & 
  \multicolumn{11}{c}{\textbf{Phase 2}}   \\
  &  &    &  \textbf{ar}  & \textbf{de} & \textbf{el} & \textbf{es} & \textbf{hi} &\textbf{ro}   &  \textbf{ru}  & \textbf{th} & \textbf{tr} & \textbf{vi} &  \textbf{zh} &  \textbf{ar}  & \textbf{de} & \textbf{el} & \textbf{es} & \textbf{hi} &\textbf{ro}   &  \textbf{ru}  & \textbf{th} & \textbf{tr} & \textbf{vi} &  \textbf{zh} \\
    \midrule
    \multirow{2}{*}{\mis} & \alp & \multirow{4}{*}{\malp} & 0.194 & 0.379 & 0.248 & 0.374 & 0.224 & 0.418 & 0.150 & 0.185 & 0.454 & 0.475 & 0.088 & 0.613 & 0.692 & 0.657 & 0.713 & 0.670 & 0.679 & 0.661 & 0.385 & 0.666 & 0.734 & 0.148 \\  
                             & Instruct &  & 0.166 & 0.568 & 0.260 & 0.510 & 0.173 & 0.508 & 0.336 & 0.210 & 0.460 & 0.502 & 0.168 & 0.369 & 0.612 & 0.253 & 0.634 & 0.450 & 0.553 & 0.555 & 0.180 & 0.532 & 0.566 & 0.089 \\  
    \multirow{2}{*}{\llm} & \alp & &  0.393 & 0.689 & 0.529 & 0.735 & 0.644 & 0.723 & 0.538 & 0.398 & 0.671 & 0.748 & 0.376 & 0.676 & 0.850 & 0.710 & 0.893 & 0.740 & 0.817 & 0.726 & 0.526 & 0.770 & 0.884 & 0.519 \\  
                              & Instruct &  & 0.659 & 0.795 & 0.702 & 0.852 & 0.715 & 0.810 & 0.609 & 0.594 & 0.728 & 0.834 & 0.533 &  0.444 & 0.580 & 0.244 & 0.657 & 0.241 & 0.586 & 0.493 & 0.092 & 0.580 & 0.558 & 0.113 \\  
    \midrule
    \multirow{2}{*}{\mis} & \orca & \multirow{4}{*}{\morca} & 0.001 & 0.010 & 0.014 & 0.001 & 0.007 & 0.009 & 0.001 & 0.006 & 0.018 & 0.001 & 0.008 & 0.639 & 0.832 & 0.570 & 0.847 & 0.601 & 0.776 & 0.771 & 0.366 & 0.734 & 0.820 & 0.113  \\  
                             & Instruct &  &  0.166 & 0.568 & 0.260 & 0.510 & 0.173 & 0.508 & 0.336 & 0.210 & 0.460 & 0.502 & 0.168 & 0.256 & 0.457 & 0.320 & 0.443 & 0.256 & 0.409 & 0.215 & 0.245 & 0.364 & 0.428 & 0.162 \\  
    \multirow{2}{*}{\llm} & \orca &  & 0.505 & 0.642 & 0.587 & 0.711 & 0.604 & 0.634 & 0.651 & 0.290 & 0.699 & 0.685 & 0.104 & 0.639 & 0.832 & 0.570 & 0.847 & 0.601 & 0.776 & 0.771 & 0.366 & 0.734 & 0.820 & 0.113 \\  
                              & Instruct &  & 0.659 & 0.795 & 0.702 & 0.852 & 0.715 & 0.810 & 0.609 & 0.594 & 0.728 & 0.834 & 0.533 & 0.654 & 0.793 & 0.703 & 0.852 & 0.718 & 0.808 & 0.606 & 0.600 & 0.729 & 0.836 & 0.540 \\  
    \bottomrule
    \end{tabular}
    \end{adjustbox}
    \caption{\texttt{XQuAD}: Language Ability results for two-phase Continual Fine-tuning (CFT).}
    \label{tab:results_xquad}
\end{sidewaystable*}

\section{Reverse Order CFT Result Analysis~\label{app::reverse_order}}
In Tables~\ref{tab:results_flip_TA} and \ref{tab:results_flip_LA}, we reverse the order of Phase~1 and Phase~2 datasets, first fine-tuning on the multilingual dataset followed by its English counterpart. For \mis\ (\malp-\alp), the average performance is 0.226, and for \llm\ (\malp-\alp), it is 0.259. Compared to the mixture setting and the \alp-\malp\ configuration (\S\ref{sec::langadapt_cft}), we observe that English ability benefits from multilingual fine-tuning in Phase~1, resulting in performance comparable to the data mixture setting. However, fine-tuning on English data in Phase~2 leads to a drastic drop in multilingual ability, yielding worse results than both the mixture setting and the two-phase setup considered in the main paper.

\section{Mitigating Strategies}\label{app::mitigating_strategies}

Here, we provide additional details on \texttt{Spectrum}~\cite{hartford2024spectrum}. We then visualize the impact of our mitigating strategies on the variance in model representations. Lastly, we ablate our findings for the Instruct-\morca\ phase-wise datasets.

\subsection{Spectrum\label{app::spectrum}}
\texttt{Spectrum}~\cite{hartford2024spectrum} is a layer-freezing technique that optimizes the fine-tuning of LLMs by selecting layers based on their signal-to-noise ratio (SNR). We use \texttt{Spectrum} as a heuristic for layer-freezing; that is, the layers identified as "important" by \texttt{Spectrum} are frozen during Phase 2 fine-tuning. A layer is important based on its signal-to-noise (SNR) ratio. In the following, we elaborate on how \texttt{Spectrum} computes SNR.

\paragraph{Marchenko-Pastur distribution.}  The Marchenko-Pastur distribution~\cite{marchenko1967distribution} is given by:
$$
    \rho(\lambda) = \frac{1}{2\pi \sigma^2} \frac{\sqrt{(\lambda_+ - \lambda)(\lambda - \lambda_-)}}{\lambda},
$$
where
$$
    \lambda_{\pm} = \sigma^2 (1 \pm \sqrt{Q})^2,
$$
and $Q = \frac{N}{M}$, with $N$ and $M$ being the dimensions of a random matrix $W$, and $\sigma^2$ representing the variance of the entries in $W$.

\paragraph{SNR.} 
Let $W\in\mathbb{R}^{N\times M}$ be the weight matrix of a given layer. The empirical spectral density of $W$ is analyzed by comparing its eigenvalue distribution of $1/N \cdot W^TW$ against the theoretical Marchenko-Pastur distribution. Deviations from this distribution indicate the presence of significant signal components. We get,

$$
    \lambda_{\pm} = \sigma^2 \left(1 \pm \sqrt{\frac{M}{N}}\right)^2,
$$
where $\lambda_\pm$ are the largest and smallest eigenvalues and $\sigma$ the standard deviation. This implies the bounds of singular values of $W$ as:
\begin{equation}\label{eqn::snr_threshold}
    \epsilon_{\pm} = \frac{1}{\sqrt{N}}\sigma \left(1 \pm \sqrt{\frac{M}{N}}\right)
\end{equation}

By evaluating how the singular values of $W$ distribute relative to $\epsilon_{\pm}$, \texttt{Spectrum} assesses the SNR of each layer, as defined next.

\paragraph{Ratio}~\cite{hartford2024spectrum}. Specifically, the SNR value of a weight matrix is,
$$
\text{SNR} = \frac{\sum_{k|\sigma_k > \epsilon} \sigma_k}{\sum_{k|\sigma_n < \epsilon} \sigma_n}
$$

Here, $\epsilon$ separates signal from noisy singular values.
Layers with singular values significantly exceeding $\epsilon_+$ have a high SNR, indicating a substantial presence of informative signal components.

\paragraph{Measuring the Ratio}~\cite{hartford2024spectrum}. Having defined all ingredients above, \texttt{Spectrum} now computes each layer's SNRs. To do this, it first computes SVD~\cite{Zhang2009} of the the layer's weight matrix, calculates the SNR and normalizes it by the highest singular value. Eq.~\ref{eqn::snr_threshold} gives the noise threshold.

Now, \texttt{Spectrum} selects layers with higher SNRs, where the number of layers selected is a hyperparameter. Similar to \citet{hartford2024spectrum}, for our experiments, we select the top-50\% of layers in each module.

\subsection{\llm\ Doesn't Show Consistent Improvement with our Mitigation Strategies}
From Table \ref{tab:results_interventions}, while both \texttt{GR} and \texttt{LF} improve on the baseline \llmi\ \malp, the gains in task and multilingual ability are not comparable to \llmi.

%
%
\begin{figure}[h]
 \centering\renewcommand{\thefigure}{\thesection\arabic{figure}}
    \centering
    \begin{minipage}{0.49\columnwidth}
        \centering
        \includegraphics[width=\textwidth]{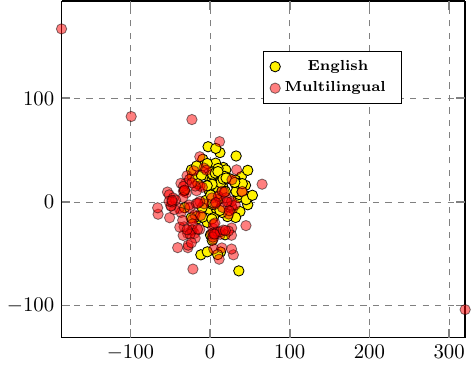}
        \caption*{(a) \mis}
    \end{minipage}
    \hfill
    \begin{minipage}{0.49\columnwidth}
        \centering
        \includegraphics[width=\textwidth]{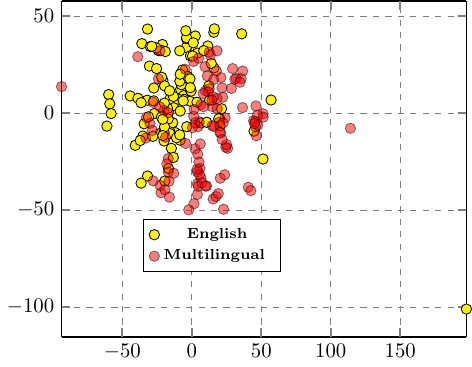}
        \caption*{(b) \llm}
    \end{minipage}
            \caption{Demonstrating extent of cross-lingual transfer in \mis\ and \llm\ on a parallel dataset prepared by subsampling \textsc{Flores}~\cite{costa2022no}. We find that the English activation cluster for \llm\ is separated from the multilingual cluster, compared to \mis. }
    \label{fig:flores}
\end{figure}
%

To understand this further, for \texttt{GR}, we investigate the cross-linguality difference between \llm\ and \mis. Like Figure~\ref{fig::tsne}, we plot t-SNEs of the mean model activations for the \mis\ and \llm\ base models on two parallel datasets, English and Multilingual. We create the parallel datasets by subsampling data from \textsc{Flores}~\cite{costa2022no}.  In Figure~\ref{fig:flores}, we see that the English activation cluster for \llm\ is separated out from multilingual cluster, compared to \mis. This suggests that \texttt{GR} may not be as effective when the model has less cross lingual ability. 
While for \texttt{LF}, we acknowledge that our method to identify the layers to freeze may not be the best and better methods to identify which layers to freeze can be a direction for future work. 

Last, but not the least, we acknowledge that \llmi\ seems to be a strong model even on multilingual benchmarks. Hence, it is also important to evaluate Phase 1 models on these benchmarks first and then decide if the Phase 2 fine-tuning step should be undertaken or not.

With regards to \llmi\ \malp\ MA results in Table~\ref{tab:results_main_MA}, we believe that this is due to lack of cross-linguality in \llmi\ and less data in \malp\ which fails to cause sufficient representation drift to improve the model's performance.

\subsection{Additional Ablations\label{app::mitigation_ablations}} We also present the impact of our mitigating strategies for the Instruct-\morca\ phase-wise datasets on \mis. Table~\ref{tab:results_interventions_ablations} presents these results.

We see that \texttt{LF\_H2} achieves moderate success, especially in maintaining the language ability for \texttt{MLQA} (0.258) and \texttt{XQUAD} (0.527). However, task ability shows some decline (e.g., \texttt{IFEval} (0.401) and \alp\ Eval (0.048)), compared to the baseline. Furthermore, \texttt{GR\_5} results in lower task ability (\texttt{IFEval} = 0.281), while \texttt{GR\_10} performs slightly better in task ability (e.g., \texttt{MMLU} = 0.483, \texttt{HellaSwag} = 0.494). Among the baselines, \texttt{ER\_10} performs similarly to the generative replay strategies, with modest improvements in task ability (e.g., \texttt{IFEval} = 0.367, \texttt{MMLU} = 0.479), but still struggles in language ability. Perhaps \texttt{LoRA} shows the best overall performance among the strategies for maintaining task ability (e.g., \texttt{IFEval} = 0.587, \texttt{MMLU} = 0.567, \texttt{HellaSwag} = 0.591) with reasonable retention of language ability (e.g., \texttt{XQUAD} = 0.354).

These results show that no single strategy is perfect, and future work may need to combine these strategies or develop new approaches to address the balance between task and language ability retention across phases.

%
\begin{table*}[t]
    \centering\renewcommand{\thetable}{\thesection\arabic{table}}
    \begin{adjustbox}{max width=\textwidth}
    \begin{tabular}{ccc|ccccc|cccc|c}
    \toprule
          \multicolumn{3}{c|}{\textbf{CFT Setup}} & \multicolumn{5}{|c|}{\textbf{Task Ability}} & \multicolumn{4}{c|}{\textbf{Language Ability}} & \textbf{Overall}  \\ 
           \midrule
       \multirow{2}{*}{\textbf{Model}}   &  \textbf{Phase 2} & \textbf{Mitigating} & \multirow{2}{*}{\textbf{IFEval}}  & \multirow{2}{*}{\textbf{\alp\ Eval}} &  \multirow{2}{*}{\textbf{MMLU}} & \multirow{2}{*}{\textbf{HellaSwag}} & \multirow{2}{*}{\textbf{Avg}} &  \multirow{2}{*}{\textbf{MLQA}} & \multirow{2}{*}{\textbf{XLSum}} & \multirow{2}{*}{\textbf{XQUAD}} & \multirow{2}{*}{\textbf{Avg}}  &  \multirow{2}{*}{\textbf{Avg}}\\  
    & \textbf{Dataset} & \textbf{Strategy} & & & & & & &  &  &  & \\
    \midrule    
    \multirow{7}{*}{\rotatebox{90}{\mis}} & \multirow{7}{*}{\morca} & -- & 0.426  & 0.060  & 0.507 & 0.509 & 0.376 & 0.155  & 0.040  & 0.323  & 0.173 &  0.275 \\
    &  & \texttt{LF\_H2} & 0.401 & 0.048 & 0.518  &  0.487   & 0.364 & 0.258 &0.060  & \textbf{0.527} & \textbf{0.282} & 0.323 \\  
    &  &  \texttt{Spectrum} &  0.442 & \textbf{0.158}  & 0.508 &  \textbf{0.616} & 0.431 & \textbf{0.387} & \textbf{0.086} & 0.201 &  {0.225} & \textbf{0.328}\\
    &  & \texttt{GR\_5} & 0.281 &   0.027&  0.478 &  0.495  & 0.320 & 0.167  & 0.042 & 0.305 & 0.171 &  0.246 \\  
    &  &  \texttt{GR\_10} & 0.305 &  0.013 & 0.483 & 0.494 & 0.324 & 0.150 &  0.038 &  0.238 & 0.142 & 0.233 \\  
    &  &  \texttt{ER\_10} &  0.367 &  0.025 &  0.479  & 0.493  & 0.341 &  0.157 & 0.042 & 0.305 & 0.168  & 0.255 \\
    &  &  \texttt{LoRA} &  \textbf{0.587} &  0.130  &   \textbf{0.567}   & 0.591 & \textbf{0.469} & 0.167   &  0.027 & 0.354  & 0.183 & 0.326 \\
    \bottomrule
    \end{tabular}
    \end{adjustbox}
   \caption{English and Multilingual Ability results for our mitigating strategies, Generative Replay (\texttt{GR\_5} \& \texttt{GR\_10}), English Replay (\texttt{ER\_10}) and Layer Freezing (\texttt{LF\_H1},  \texttt{LF\_H2} \& \texttt{Spectrum}). We use \texttt{LoRA}~\cite{hu2022lora} as a baseline strategy. For \texttt{ER\_10}, we use the English dataset used in \texttt{GR} with original responses. \textit{The Phase 1 dataset is Instruct for each row.} The first row provides \mis\ numbers for Instruct-\morca\ (from Table~\ref{tab:results_main_ablation}).
    }
    \label{tab:results_interventions_ablations}
\end{table*}

\subsection{Compute Analysis\label{app::compute_analysis}}
The computational overhead of replay arises from an increase in the effective training set size . In \texttt{ER\_10}, each epoch includes the original Phase~2 data along with an additional 10\% replay buffer from Phase~1, resulting in a proportional increase in the number of training tokens and hence compute. Similarly, \texttt{GR\_5} augments each epoch with 5\% generated replay data, leading to a smaller overhead.

In contrast, layer freezing (\texttt{LF}) significantly reduces computation by updating only a subset of model parameters. By freezing 50\% of the layers, \texttt{LF} yields a corresponding reduction in training cost while still achieving a reasonable trade-off between English retention and multilingual adaptation.

\section{Resources Used}
\label{sec:resources_used}
We conduct all experiments on 4 NVIDIA A100 GPUs (80GB each) with a 96-core AMD CPU. A single fine-tuning run on \malp\ takes approximately 4 hours, while training on \morca\ requires around 12 hours. The models used in our experiments, along with their corresponding checkpoints and licenses, are listed below:
\begin{itemize}
    \item \llm: \url{https://huggingface.co/meta-llama/Meta-Llama-3-8B} \hfill {\bf License:} LLaMA 3
    \item \mis: \url{https://huggingface.co/mistralai/Mistral-7B-v0.1} \hfill {\bf License:} Apache-2.0
\end{itemize}

\section{Existing Mitigating Strategies}
\label{sec:cft_comparison}
We now discuss why MAD-X~\cite{pfeiffer2020madxadapterbasedframeworkmultitask} and other PEFT-based methods~\cite{badola-etal-2023-parameter} are not well-suited to our setting.

\begin{itemize}[leftmargin=*]
    \item \textbf{Parameter-Efficient Fine-tuning for Robust Continual Multilingual Learning}~\cite{badola-etal-2023-parameter}: This work considers a fixed downstream task (text classification) that is incrementally extended to new languages, mitigating forgetting via task- and language-specific adapters. Their approach assumes continued access to prior task data and the ability to maintain multiple adapter checkpoints. In contrast, our setting involves end-to-end instruction fine-tuning over a heterogeneous mixture of tasks. We do not assume access to Phase~1 data during Phase~2, nor do we maintain multiple parameter sets due to the memory overhead of storing additional model weights (\S\ref{sec::exp_setup}).

    \item \textbf{MAD-X}~\cite{pfeiffer2020madxadapterbasedframeworkmultitask}: MAD-X introduces bottleneck adapters within encoder-only or encoder--decoder Transformer architectures to enable cross-lingual transfer. However, recent work~\cite{zhao2024adamergexcrosslingualtransferlarge} shows that such adapters do not integrate cleanly with decoder-only LLMs (e.g., Mistral-7B, LLaMA-3-8B), which form the basis of our experiments. As our models lack an encoder stack, the routing and merging mechanisms central to MAD-X are not directly applicable.

    \item \textbf{Preserving Cross-Linguality via Continual Learning}~\cite{liu-etal-2021-preserving}: This method extends Gradient Episodic Memory by storing exemplars (or their gradients) from previous phases and replaying them during training. Our setup (\S\ref{sec::exp_setup}) explicitly disallows (i) reusing Phase~1 data and (ii) maintaining Phase~1 representations or weights during Phase~2, as these incur significant memory overhead and may violate practical data-sharing constraints.
\end{itemize}



\end{document}